\documentclass[compsoc,journal]{IEEEtran}
\usepackage{graphicx}
\usepackage[justification=justified]{caption}
\usepackage{ragged2e}
\usepackage{enumerate}
\usepackage{url}
\usepackage{multirow}
\usepackage{amsmath}
\usepackage{bm}
\usepackage{hyperref}
\usepackage{diagbox}
\usepackage{gensymb}
\usepackage{makecell}
\usepackage{subfig}

\hypersetup{
    colorlinks=true,
    linkcolor=black,
    filecolor=black,
    urlcolor=black,
    citecolor=black,
}

\ifCLASSOPTIONcompsoc
  \usepackage[nocompress]{cite}
\else
  \usepackage{cite}
\fi

\begin{document}

\title{A Screen-Shooting Resilient Document Image Watermarking Scheme using Deep Neural Network}

\author{
	Sulong~Ge,
	Zhihua~Xia,
	Yao~Tong,
	Jian~Weng, 
	and Jianan~Liu % <-this % stops a space
	\IEEEcompsocitemizethanks{
		\IEEEcompsocthanksitem Sulong Ge is with Engineering Research Center of Digital Forensics, Ministry of Education, School of Computer and Software, Nanjing University of Information Science \& Technology, Nanjing, 210044, China.
		\IEEEcompsocthanksitem Zhihua Xia, Jian Weng, and Jianan Liu are with College of Cyber Security, Jinan University, Guangzhou, 510632, China. Zhihua Xia is also with Engineering Research Center of Digital Forensics, Nanjing University of Information Science \& Technology, Nanjing, 210044, China.
		\IEEEcompsocthanksitem Zhihua Xia and Jian Weng are the corresponding authors. e-mail: xia\_zhihua@163.com, cryptjweng@gmail.com. 
		\IEEEcompsocthanksitem The code of our scheme can be found at https://github.com/gslxr/Screen-Shooting-Resilient-Document-Image-Watermarking \protect\\ 
	}% <-this % stops a space
}

\markboth{Journal of \LaTeX\ Class Files,~Vol.~14, No.~08, August~2015}%
{Ge \MakeLowercase{\textit{et al.}}:Screen-Shooting Resilient Document Image Watermarking using Deep Neural Network}

\IEEEtitleabstractindextext{
	\begin{abstract}
		With the advent of the screen-reading era, the confidential documents displayed on the screen can be easily captured by a camera without leaving any traces. Thus, this paper proposes a novel screen-shooting resilient watermarking scheme for document image using deep neural network. By applying this scheme, when the watermarked image is displayed on the screen and captured by a camera, the watermark can be still extracted from the captured photographs. Specifically, our scheme is an end-to-end neural network with an encoder to embed watermark and a decoder to extract watermark. During the training process, a distortion layer between encoder and decoder is added to simulate the distortions introduced by screen-shooting process in real scenes, such as camera distortion, shooting distortion, light source distortion. Besides, an embedding strength adjustment strategy is designed to improve the visual quality of the watermarked image with little loss of extraction accuracy. The experimental results show that the scheme has higher robustness and visual quality than other three recent state-of-the-arts. Specially, even if the shooting distances and angles are in extreme, our scheme can also obtain high extraction accuracy.
	\end{abstract}

	}
\maketitle

\IEEEdisplaynontitleabstractindextext
\IEEEpeerreviewmaketitle

\section{Introduction}\label{sec:introduction}

\begin{figure}[htbp]
	\includegraphics[width=\linewidth]{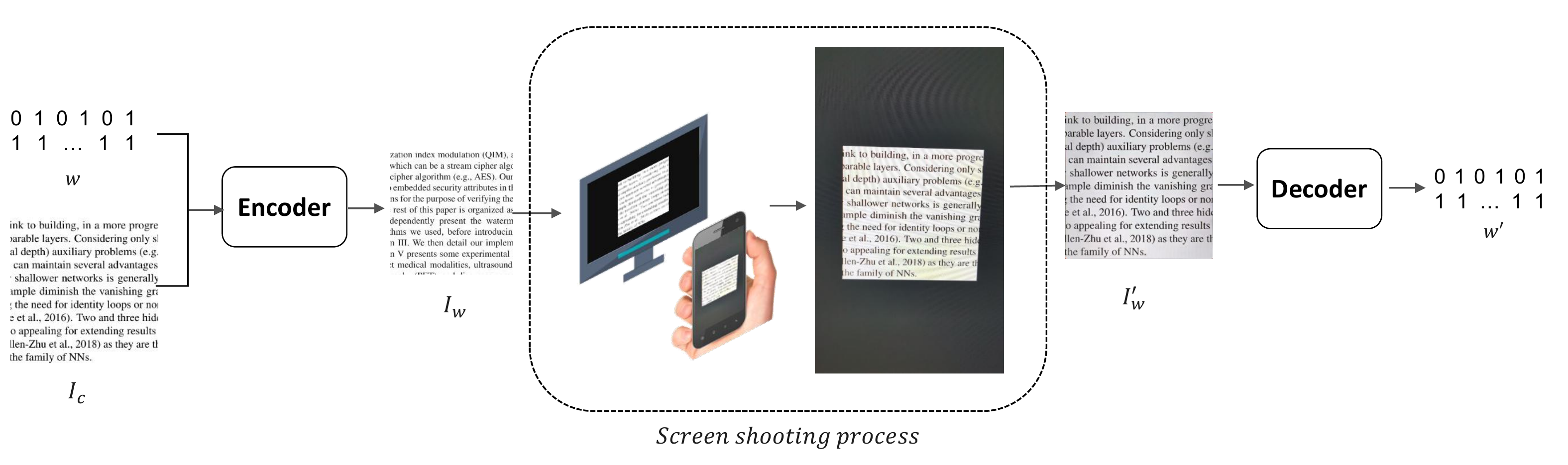}
	\caption{
		A framework of the screen-shooting resilient image watermarking.}
	\label{fig:watermarking_framework}
 \end{figure}

 With the popularity of smart phones, taking photographs has become the most simple and efficient way of information transmission, which brings a threat to information security. The commercialization process generates lots of valuable secret files, anyone who has access to the files, can simply steal information by taking photographs without leaving any records \cite{gugelmann2018screen}. In addition, the screen-shooting process is difficult to prohibit from the outside, so it is very important to design a screen-shooting resilient (SSR) document watermarking scheme to solve this problem. As shown in Fig.~\ref{fig:watermarking_framework}, we can embed watermark such as identity information in the document, so when the document after the screen-shooting process, we can also extract the identification from the captured photographs.

 Screen-shooting process is the process of capturing images displayed on the screen with a camera. It can be seen as one of the called cross-media information transmission processes, and the other are print-scanning process and print-camera process \cite{fang2018screen}. In previous years, the print-scanning resilient (PSR) watermarking schemes and the print-camera resilient (PCR) watermarking schemes have been extensively studied. Print-scanning process is the process of printing image on paper and then scanning through a scanner. Rotation, scaling, and translation (RST) are the common distortions during the process. So far, PSR watermarking scheme can be broadly divided into two categories that are template-based methods \cite{pereira2000robust, kang2003dwt, pramila2007multiple} and transform invariant domain-based methods \cite{he2005practical, solanki2006print, kang2010efficient, amiri2014robust}. Besides, Print-camera process can be seen as an addition to the print-scanning process, which means printing image on paper and
 taking a photograph of the printed image. The watermark must be robust to more distortions, such as camera distortion, shooting distortion and so on. PCR watermarking schemes can be roughly divided into three categories. One is template-additive based methods \cite{nakamura2004fast, kim2006image, pramila2012toward}. The other two are transformed invariant domain-based methods \cite{delgado2013digital, gourrame2016robust, liang2019robust} which is developed from PSR watermarking scheme, and the DNN-based (deep neural network) methods \cite{tancik2020stegastamp, jia2020rihoop}.

 With the advent of screen-reading era, the screen-shooting process is becoming common, and using this way to leak files is the most difficult problem at this stage. Compared with the print-scanning process and the print-camera process, the watermarked image undergoes a series of analog-to-digital (AD) and digital-to-analog (DA) conversion processes in the screen-shooting process, which will be manifested as a combination of strong distortions \cite{fridrich2009digital}. Therefore, it becomes impractical to design watermarking algorithms by seeking transformed invariant domain, like \cite{fang2018screen, fang2019camera, liang2019robust}, we need to analyze all the distortions in the screen-shooting process and design corresponding compensation operations for the distortions. With the development of deep learning, deep learning instead of artificial design has become the mainstream of natural image watermarking algorithms, and has achieved good results in robustness and visual quality. However, according to the peculiarity of text content, these algorithms cannot be applied to document images without adjustments. In this paper, we propose a screen-shooting resilient document image watermarking scheme using deep neural network. The main contributions of this paper are listed as follows: 

 \begin{itemize} 
 	\item We propose a document image watermarking scheme which is resilient to the screen-shooting process. Our scheme outperforms the recent state-of-the-arts in both the visual quality and robustness. 
 	
 	\item During the training process, a distortion layer is designed to simulate the distortions caused by the screen-shooting process, including the camera distortion, shooting distortion, light source distortion, and etc. Such design is quite helpful in improving the robustness.
 	
 	\item Redundancy space in model parameters during the training process provides possibility for image co-encoding. An embedding strength adjustment strategy is proposed to improve the visual quality of the watermarked image, which almost has no effect on the extraction accuracy.
 
 \end{itemize}

 The rest of this paper is organized as follows. The Section~\ref{sec:releted_work} introduces the related works. The screen-shooting distortions is analyzed in Section~\ref{sec:ssr_analysis}. In the Section~\ref{sec:proposed_scheme} shows the details of the proposed scheme. The Section~\ref{sec:experiment_results} discusses the experimental results and analysis. The Section~\ref{sec:conclusion} draws the conclusion.

\section{Related works}\label{sec:releted_work}

 In this section, we summarize the image watermarking schemes about the cross-media information transmission process in recent years, including the print-scanning resilient (PSR) schemes, the print-camera resilient (PCR) schemes, and the screen-shooting resilient (SSR) schemes.

\subsection{Print-scanning resilient watermarking}

 Print-scanning process means printing image on paper and then scanning through a scanner. PSR schemes can be broadly divided into two categories: the template-based methods and the transform invariant domain-based methods.

 \textbf{The template-based methods.} This kind of methods design template for watermark to resist print-camera process. Pereira and Pun \cite{pereira2000robust} proposed a templete watermarking scheme using discrete Fourier transform (DFT). The authors proved that watermark embedding into the DFT domain can resist affine transformation. Based on this, Kang \emph{et al.} \cite{kang2003dwt} proposed a watermarking using discrete wavelet transform (DWT) and DFT. Watermark is embedded in the coefficients of the LL subband in the DWT domain. To resist affine transformation, a template is embedded in the middle frequency components in the DFT domain. Pramila \emph{et al.} \cite{pramila2007multiple} proposed a watermarking scheme in multiple domains. To resist rotation and scaling after print-scan process, a circular template is embedded in the DFT domain. To resist translations, another template is embedded in spatial domain. To achieve high robustness and visual quality, the watermark is embedded in the wavelet domain. 

 \textbf{The transform invariant domain-based methods.} This kind of methods design watermarking schemes in the transform invariant domain that can resist the print-camera process. He and Sun \cite{he2005practical} proposed a watermarking scheme using DFT. The cover is divided into the non-overlapping blocks, in which watermark is embedded into the middle frequency DFT coefficients block by block. The watermark extraction is an inverse procedure of the watermark embedding process. Solanki \emph{et al.} \cite{solanki2006print} proposed a watermarking scheme using the analytical modeling of the print-scan process, in which geometric transformations, nonlinear effects, and colored noise are main components. Watermark is embedded into high-magnitude of low-frequency DFT coefficients. Besides, A novel approach was proposed for estimating the rotation that an image might undergo during the scanning process. Kang \emph{et al.} \cite{kang2010efficient} proposed a watermarking scheme using uniform log polar mapping (ULPM), which is resilient to both geometric distortion and the print-scan process. To obtain a discrete log polar point, the authors apply ULPM to the frequency index in the Cartesian system. Then the watermark is embedded in the corresponding 2D-DFT coefficients in the Cartesian system. Amiri and Jamzad \cite{amiri2014robust} proposed a watermarking scheme using DWT and discrete cosine transform (DCT). The authors apply Two-dimensional DWT to the cover for obtaining the mid-frequency subbands, then apply DCT to the selected subbands for embedding the watermark. Besides, a Genetic Algorithm is used to achieve image quality.

\subsection{Print-camera resilient watermarking}

 Print-camera process means printing image on paper and then taking a photograph of the printed image. PCR schemes can be roughly divided into two categories: the template-based methods, the transformed invariant domain-based methods and the DNN-based methods. 
 
 \textbf{The template-based methods.} This kind of methods design templates for watermarks to resist print-camera process. Nakamura \emph{et al.} \cite{nakamura2004fast} proposed a fast watermark detection scheme from a captured image. The method consists of two processes, one is to correct geometric distortion of the captured image, and the other is to detect watermark from the rectified image. Kim \emph{et al.} \cite{kim2006image} proposed an watermarking scheme for the print-camera process. They embed watermark into the spatial domain of a color image, print out using a printer, and extract the watermark from the image captured by a digital camera. Pramila \emph{et al.} \cite{pramila2012toward} proposed a directed periodic pattern-based watermarking scheme. This scheme generate a periodic template and embed the watermark by modulating the direction of the template. At the extraction side, the Hough transform is used to detect the angle of template.

 \textbf{The transform invariant domain-based methods.} This kind of methods design watermarking schemes in the transform invariant domain that can resist the print-camera process. Delgado-Guillen \emph{et al.} \cite{delgado2013digital} proposed a watermarking scheme for mobile platforms. The embedding and extracting algorithms are based on the scheme proposed by Kang \emph{et al.} \cite{kang2010efficient}. Gourrame \emph{et al.} \cite{gourrame2016robust} proposed a watermarking scheme using DFT. The watermark is embedded into the luminance channel by adjusting the coefficient of the DFT. For the perspective distortions, the 3D rotation of the image and camera position are simulated simultaneously. Liang and Wang \cite{liang2019robust} proposed a watermarking scheme using DCT and scale-invariant feature transform (SIFT). This method embed watermarks into middle frequency of the DCT coefficients. The SIFT is used to revert the geometric distortions and correct blurred pixel values.

 \textbf{The DNN-based methods.} This kind of methods train network to simulate the print-camera process. Tancik \emph{et al.} \cite{tancik2020stegastamp} proposed an end-to-end watermarking scheme for print-camera process. The authors use a set of differentiable image perturbations (Perspective Warp, Color Manipulation, and JPEG Compression) between the encoder and decoder to approximate the distortions in print-camera process. It proved that this method has a quite high robustness to the print-camera distortions. On this basis, Jia \emph{et al.} \cite{jia2020rihoop} proposed a watermarking scheme using differentiable 3D rendering and just noticeable difference (JND) loss. To make the watermark resilient to print-camera process during training process, this method introduce a distortion network (DL) using differentiable 3D rendering to augment the encoded images, in which almost all distortions in print-camera process are simulated. Besides, the JND-based loss is helpful to maintain image quality.

\subsection{Screen-shooting resilient watermarking}

 Screen-shooting process means capturing images displayed on the screen with a camera. SSR schemes can be roughly divided into two categories: the transformed invariant domain-based methods and the DNN-based methods.

 \textbf{The transform invariant domain-based methods.} This kind of methods design watermarking schemes in the transform invariant domain that can resist the screen-shooting process. Fang \emph{et al.} \cite{fang2018screen} proposed a watermarking scheme using an intensity-based scale-invariant feature transform (I-SIFT) algorithm and DCT transform. It proved that this method has good robustness to the screen-shooting process and visual quality. However, compared with natural images, the simple texture and color of the document would result in the weak strength of I-SIFT key point, so that the watermark region cannot be accurately located. Subsequently, Fang \emph{et al.} \cite{fang2019camera} proposed the flip-based watermarking scheme to better synchronize the watermark region. Li \emph{et al.}\cite{li2021screen} proposed a watermarking scheme using learned invariant keypoints, quaternion discrete Fourier transform (QDFT), and tensor decomposition (TD).They apply feature regions filtering model to SuperPoint (FRFS) to locate the embedding regions, then the watermarks are embedded by using the QDFT and TD algorithm. To enhance robustness, the watermarks are repeatedly embedded into different regions in an image. 

 \textbf{The DNN-based methods.} This kind of methods train network to simulate the screen-shooting process. Wengrowski and Dana \cite{wengrowski2019light} proposed an end-to-end watermarking for the screen-shooting process. To learn the camera-display transfer function (CDTF), they introduce a dataset (Camera-Display) of 1,000,000 camera-captured images collected from 25 camera-display pairs. This method significantly outperformed existing methods, even when camera and display were at high perspective angles, but seriously driven by datasets \cite{zhang2021brief}. Zhang \emph{et al.} \cite{zhang2020udh} proposed a watermarking shceme with Universal Deep Hiding (UDH). The authors proved that only adding a perspective transformation layer between the encoder and decoder, which can resist the distortions introduced by screen-shooting process to a certain extent. Unlike the above schemes, Fang \emph{et al.} \cite{fang2020deep} proposed a watermarking scheme using DNN. The scheme is not end-to-end, the watermark template and the locating template are generated by template generation scheme, and then embedded into the red or blue channel of the cover respectively. Watermark extraction is designed by DNN, which is composed of an enhancing network and a classification network. Subsequently, Fang \emph{et al.} \cite{fang2021tera} also proposed a watermarking scheme named "TERA". The scheme is like \cite{fang2020deep}, watermark embedding depends on manual design named superposition-based embedding scheme and watermark extraction is designed by deep neural network (DNN). Extraction network is composed of an enhancing network, an attention-guided network \cite{vaswani2017attention} and a regression network.

\section{Screen-shooting distortion analysis}\label{sec:ssr_analysis}

 Combined with the analysis of Fang \emph{et al.} \cite{fang2018screen} and Jia \emph{et al.} \cite{jia2020rihoop}, we summarize the distortions in the screen-shooting process into three aspects: camera distortion, shooting distortion and light source distortion.

\subsection{Camera Distortion}\label{sec:camera_distortion}

 Camera distortion mainly comes from the distortion when taking photographs with the camera, and different types of cameras produce different camera distortions. The sources of these distortions are generally classified into three categories: 1) the distortion caused by electronic components inside the camera system. 2) the distortion from camera parameters. 3) the distortion caused by image compression. The electronic components in the camera system can introduce a series of noises in the imaging process, including photon noise, dark noise and lens noise \cite{hasinoff2014photon}. The camera parameters mainly affect the focusing in the imaging process, and the defocusing phenomenon will lead to blurred images. Photographs are usually saved in a lossy format (JPEG compression) after being shot, and JPEG compression will bring image loss to the captured photographs.

\subsection{Shooting Distortion}\label{sec:shooting_distortion}

 Due to the diversity of camera angles and the distances between the object and the camera lens, when we use a camera to capture the image displayed on the screen, the image will be firstly affected by perspective deformation from the original 3D space to the 2D plane, and then followed by rotation, scaling, and translation distortions. 

\subsection{Light Source Distortion}\label{sec:light_source_distortion}

 Our ability to see the world depends on the reflection of light from the surface of any object. Therefore, lighting conditions are the most important and complex environmental factors in the camera imaging process. For light source distortion, in addition to the external light, the screen itself is also a light source. The unevenness of the screen light source will produce brightness, contrast and saturation distortions \cite{jia2020rihoop}.

 \begin{figure*}[ht]
	\centering
	\includegraphics[width=\linewidth]{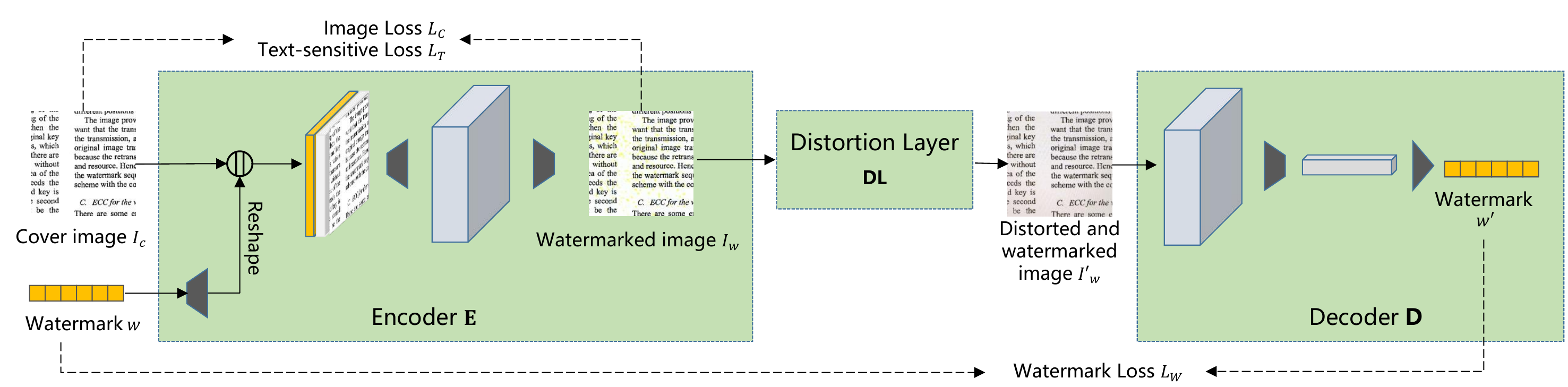}
	\caption{ 
		The overview of the proposed scheme.}
	\label{fig:overview}
 \end{figure*}

\section{proposed scheme}\label{sec:proposed_scheme}

 In this section, we firstly give an overview of our scheme. Then, the encoder, distortion layer, decoder, and loss function are specified. 

\subsection{The overview of the proposed scheme}\label{sec:overview_proposed_scheme}

 As shown in Fig.~\ref{fig:overview}, the overall architecture of our scheme is introduced, which is an end-to-end training pipeline composed of three components, namely encoder (E), distortion layer (DL) and decoder (D). In the training stage, the watermark $w$ is expanded, reshaped, and concatenated with the cover image $I_c$. Then, combination of $I_c$ and $w$ are encoded to generate the watermarked $I_w$. Next, $I_w$ is input into the distortion layer which simulates the screen-shooting process, generating the distorted and watermarked image $I'_w$. Finally, $I'_w$ is input into the Decoder \textbf{D} to extract the watermark $w'$. After training, the Encoder \textbf{E} is used to embed the watermark and the Decoder \textbf{D} is used to extract the watermark.

\subsection{Encoder}\label{sec:encoder}

 \begin{figure*}[htbp]
	\centering
	\includegraphics[width=\linewidth]{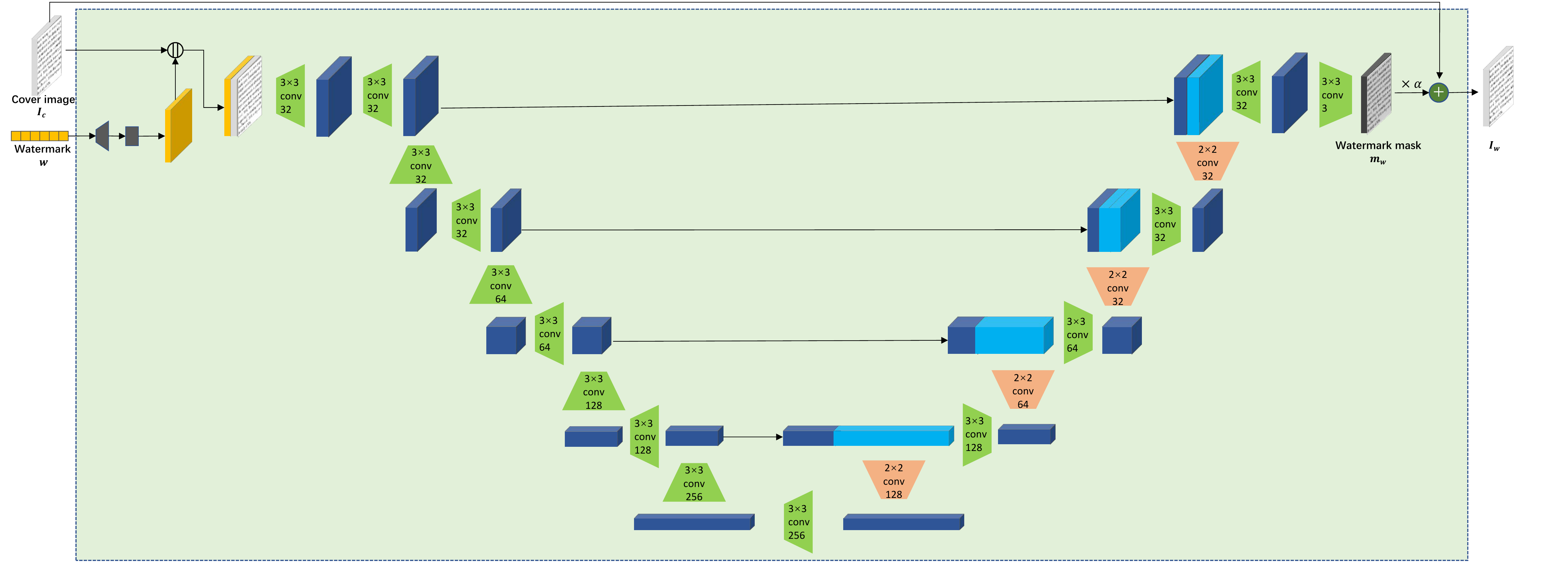}
	\caption{
		The structure of the Encoder \textbf{E}.}
	\label{fig:encoder}
 \end{figure*}

 Encoder \textbf{E} is a network trained to embed the watermark into the cover image as illustrated in Fig.~\ref{fig:encoder}. Since the U-Net \cite{ronneberger2015u} like architectures has proved its capability in many image-to-image translation tasks, we use the U-Net style architecture as our \textbf{E}, which receives a six-channel input tensor and outputs a three-channel mask image. The watermark is represented as a binary bit string with L bits. To enhance the robustness to distortions in training stage, the watermark is expanded by a fully connected layer to bring in redundancy, and then up-sampled to the same size as the cover image. The expanded watermark with the cover are concatenated and input to \textbf{E}, and finally outputs a mask image which represent the watermark. Therefore, the watermarked image is the combination of cover image and mask image, that is

 \begin{equation} 
	\label{equ:addition} 
		I_w = I_c + \alpha \cdot m_w,
 \end{equation}

 \noindent where $\alpha$ is an embedding strength factor.

 \subsection{Distortion layer}\label{sec:distortion_layer}

 \begin{figure}[htbp]
	\includegraphics[width=\linewidth]{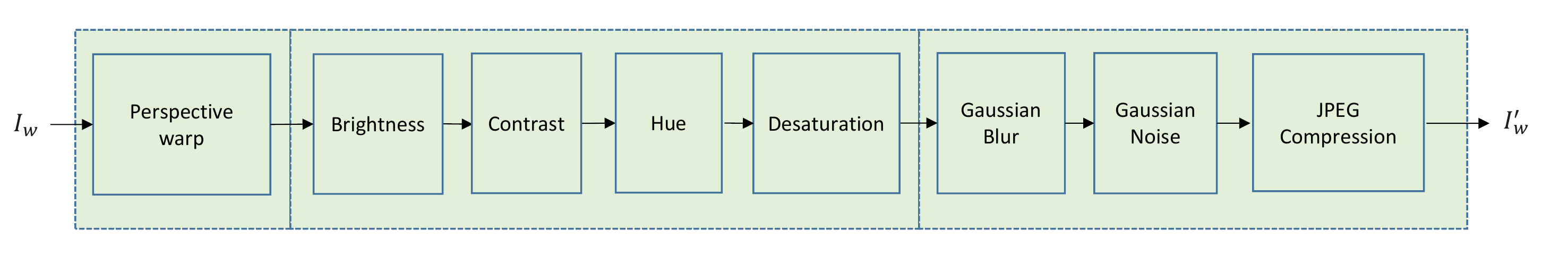}
	\caption{
		The structure of Distortion Layer \textbf{DL}.}
	\label{fig:distortion_layer}
 \end{figure}

 In the Section~\ref{sec:ssr_analysis}, we divide the screen-shooting distortions into camera distortion, shooting distortion and light source distortion. To resist the distortions introduced in the screen-shooting process, we can simulate it in the training process. So, the process is added to the output of the Encoder \textbf{E} as Distortion Layer \textbf{DL}. 

 \textbf{Camera distortion}. Electronic components in the camera system introduce multiple noises in the imaging process. To simulate the random noises, we use a random Gaussian noise ($\alpha$ = U[0, 0.02]). To simulate defocus blur, we apply Gaussian kernel with width from 1 to 3 pixels and variance from 0.01 to 1. To simulate motion blur, we use a straight line blur kernel with a width from 3 to 7 pixels and a random angle from 0 to 2 $\pi$. To compensate the JPEG compression loss, as the quantization of JPEG compression is not differentiable, we use the trick proposed by Shin and Song \cite{shin2017jpeg}, which is a differentiable operation to approximate the quantization step close to 0.

 \textbf{Shooting distortion}. The camera angles and distances make the camera lens deviate from the image plane, resulting in the perspective deformation. To simulate perspective deformation, we randomly perturb the four corners of the watermarked image in a fixed range (up to $\pm$ 40 pixels, i.e. $\pm$ 10\%), then applying homography to restore the watermarked image into the new locations.
 
 \textbf{Light source distortion}. Compared with the entire RGB color space, the color area of the display is very limited. The camera system uses many manipulations to modify its output, such as exposure settings, color correction matrix, and white balance. For convenience, we use a series of random affine color transformations to approximate these pixel distortions, including global contrast, brightness, and hue adjustments, which can achieve a similar effect to light reflection. Please note that, the details can be found from the link: https://github.com/gslxr/Screen-Shooting-Resilient-Document-Image-Watermarking.
 
\subsection{Decoder}\label{sec:decoder}

 \begin{figure*}[htbp]
	\includegraphics[width=\linewidth]{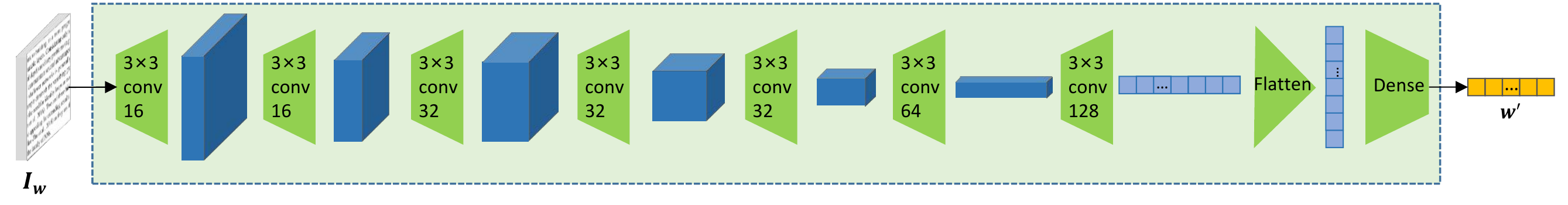}
	\caption{
		The structure of the Decoder \textbf{D}.}
	\label{fig:decoder}
 \end{figure*}

 Decoder \textbf{D} is a network trained to extract the watermark from the distorted and watermarked image $I'_w$ as illustrated in Fig.~\ref{fig:decoder}. \textbf{D} is composed of a series of convolution layers, a flatten layer, and a fully connected layer. The output of the Distortion Layer \textbf{DL} is input into \textbf{D} to extract the watermark. Finally, the sigmoid function is used to activate the output of the fully connected layer. 
 
\subsection{Loss function}\label{sec:loss_function}

 In this section, we will give the loss function, which consists of three parts: image loss, text sensitive loss and watermark loss.
 
 \textbf{Image loss}. Image loss is used to make the watermarked image $I_w$ and the cover $I_c$ look the same. We consider the image in YUV color space and try to make little change on the Y component as the human eyes are more sensitive to it. The image loss is designed as follows,

	\begin{equation}
	\begin{split}
		L_{I} 	&= MSE(I_{w}^{Y},I_{c}^{Y}) \times s_{Y} \\
		 		&+ MSE(I_{w}^{U},I_{c}^{U}) \times s_{U} \\
		     	&+ MSE(I_{w}^{V},I_{c}^{V}) \times s_{V},
	\end{split} 
	\end{equation}
 \noindent where $MSE$ denotes the mean squared error, and $s_{Y}$, $s_{U}$ and $s_{V}$ are the weights for YUV channels. 
 
 \textbf{Text-sensitive loss}. Readers will pay more attention to the text content in the reading process, so the modification on characters can be more conspicuous than on background \cite{zhao2016loss}. Thus, text-sensitive loss is designed to restrain the modification on characters as follows,
	\begin{equation}
 	\begin{split}
		L_{T} 	&= \mid I_{w}^R - I_{c}^R \mid \cdot \tilde{I_{c}^R} \times s_{R}\\ 
				&+ \mid I_{w}^G - I_{c}^G \mid \cdot \tilde{I_{c}^G} \times s_{G}\\
				&+ \mid I_{w}^B - I_{c}^B \mid \cdot \tilde{I_{c}^B} \times s_{B},
	\end{split} 
	\end{equation}

 \noindent where $\tilde{I_{c}^*}=\frac{255-I_{c}^*}{255}$ assigns larger punishment to the text content, and $s_{*}$ denotes the weights for different color components, $* \in \{R,G,B\}$.

 \textbf{Watermark loss}. Watermark loss is used to minimize the difference between the extracted and the original watermark. Binary cross entropy function is used for it as follows,
 
	 \begin{equation} 
	 	L_{W} = -\sum_{i=1}^{N}{(w_i \cdot \log(w'_i) + (1 - w_i) \cdot \log(1 - w'_i))}, 
	 \end{equation} 
	 
 \noindent where $w_i$ refers to the original watermark, $w'_i$ denotes the extracted watermark, and $N$ is the watermark length.

 Finally, the total training loss can be calculated as
	\begin{equation} 
		L_{total} = \lambda_{I}L_{I} + \lambda_{T}L_{T} + \lambda_{W}L_{W}. 
	\end{equation}
 \noindent where $\lambda_{I}$, $\lambda_{T}$, and $\lambda_{W}$ are weight factors.

\section{Experimental results and analysis}\label{sec:experiment_results}
 In this section, we show the implementation details and experimental results. Besides, an embedding strength adjustment strategy is discussed.

\subsection{Implementation Details}\label{sec:implentation_details}

 We respectively select the first 100,000 document images from DocImgEN and DocImgCN training sets \cite{ge2022robust} as our training sets, and then selected the first 100 document images from DocImgEN and DocImgCN test sets as our test sets. In the experiment, the size of the cover is 400 $\times$ 400 pixels, and the length of the watermark is 100 bits which is random bit string. 
 
 In the training stage, we feed the document images and watermark into the Encoder \textbf{E} by batches, and the size of each batch is 4. To accelerate the convergence of model parameters, we set the learning rate to 0.0001 and use the Adam optimizer \cite{kingma2014adam} to optimize the model parameters. An NVIDIA 1080Ti GPU is used as the training environment for model training. 

 In addition, we find that even with many rounds of training, the decoding accuracy is still difficult to increase. So, we froze the encoder at the first 3000 iterations, just training the decoder and distortion layer. Besides, to make the Decoder \textbf{D} gradually adapt to the distortions, the parameters $\lambda_{I}$, $\lambda_{T}$, and $\lambda_{W}$ are set to be 0 at the beginning, and increase linearly to 1.5, 1.5, and 2.0 at the first 15,000 iterations. Please note that, the other parameters can be found from the link: https://github.com/gslxr/Screen-Shooting-Resilient-Document-Image-Watermarking.

\subsection{The quality of the watermarked document image}\label{sec:watermarked_document_image_quality}
 In this section, to evaluate the quality of watermarked images, we select Peak Signal to Noise Ratio (PSNR) and Structural Similarity Metric (SSIM) \cite{wang2004image} as our metrics. Considering that readers may be more sensitive to the characters while reading the documents, we use the text-sensitive loss to ensure less modification on text pixels. Here we use the \textbf{C}hange Intensity \textbf{P}er Text-\textbf{P}ixel (CPP) designed in \cite{ge2022robust} to evaluate the modification on text pixels. The CPP metric is designed as follows,
 
 \begin{equation} 
	\begin{split}
		CPP &= \frac{\sum_{i=1}^{n_t}{\mid I_{c}^{tR}(i) - I_{w}^{tR}(i) \mid}}{n_t}\\ 
			&+ \frac{\sum_{i=1}^{n_t}{\mid I_{c}^{tG}(i) - I_{w}^{tG}(i) \mid}}{n_t}\\
			&+ \frac{\sum_{i=1}^{n_t}{\mid I_{c}^{tB}(i) - I_{w}^{tB}(i) \mid}}{n_t},
	\end{split} 
 \end{equation}

 \noindent where $I_{\#}^{t*}, \#\in {c,w}, *\in \{R,G,B\} $ refers to the set of text pixels in the cover and watermarked images, and $n_t$ denotes the total number of the text pixels. 

\begin{figure}[htbp]
	\includegraphics[width=\linewidth]{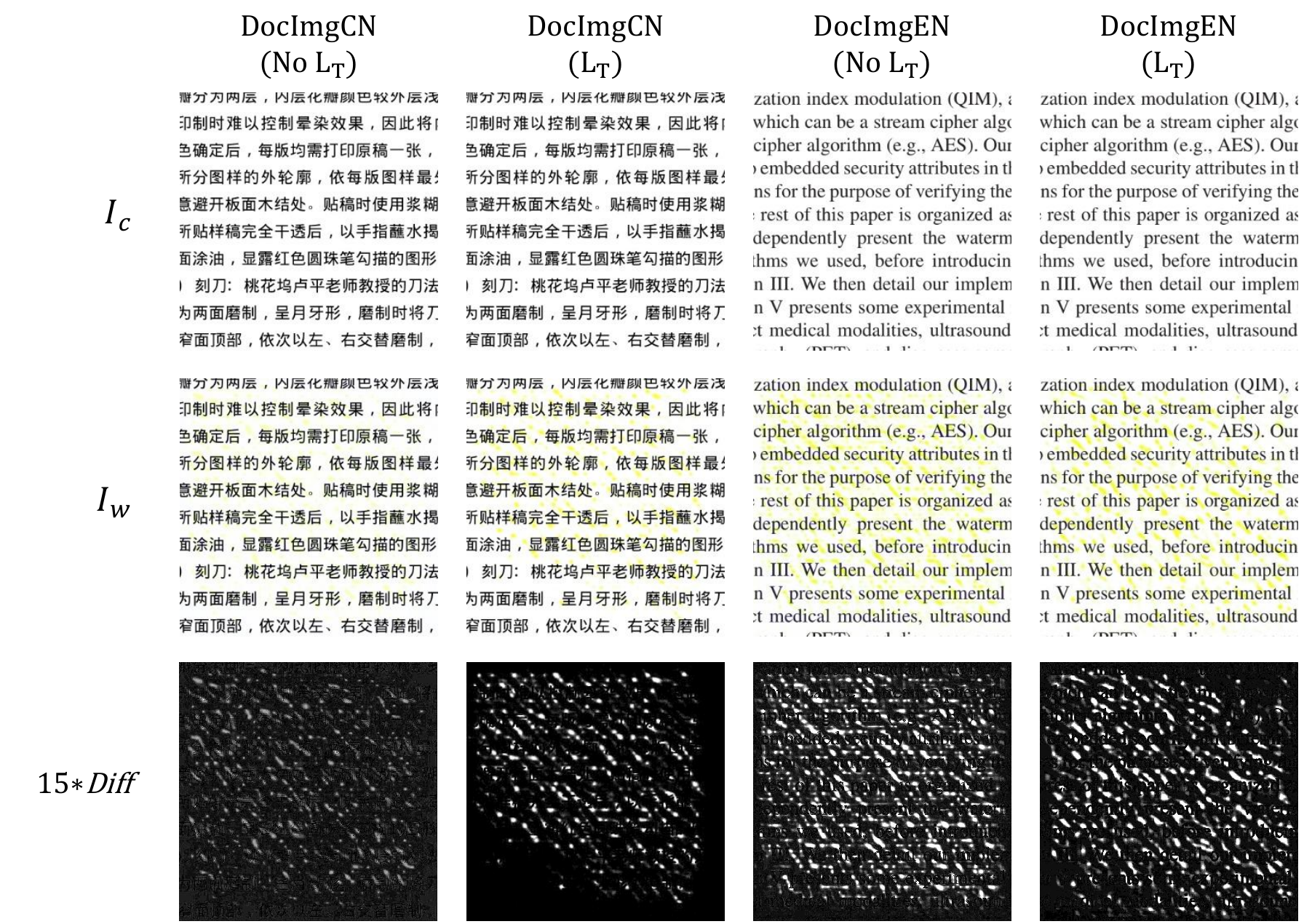}
	\caption{
		Example pairs of cover and watermarked images and their difference with the text-sensitive loss $L_{T}$ or not.}
	\label{fig:imagequality}
 \end{figure}

 The PSNR, SSIM and CPP values are listed in Table~\ref{Table:ImgQuality}, and some example pairs of cover and watermarked images are shown in Fig.~\ref{fig:imagequality}. As listed in Table~\ref{Table:ImgQuality}, the incorporation of text-sensitive loss $L_T$ make little influence on PSNR and SSIM values but decrease CPP value by up to 53.84\% averagely. It indicates that $L_T$ loss decreases the modification on text pixels and has little effect on image quality of the watermarked image. 

 \begin{table}[htbp]
	\caption{
		PSNR, SSIM, and CPP of our scheme with the text-sensitive loss $L_{T}$ or not.}
	\centering
	\begin{tabular}{l|ccc}
		\hline \hline
		Schemes							& PSNR (dB)			& SSIM				& CPP \\ \hline
		On DocImgEN without $L_{T}$		& 32.90				& \textbf{0.916}	& 15.55 \\
		On DocImgEN with $L_{T}$		& \textbf{34.10}	& 0.914				& \textbf{6.88} \\ \hline
		On DocImgCN without $L_{T}$		& \textbf{35.70}	& \textbf{0.946}	& 11.40 \\
		On DocImgCN with $L_{T}$		& 33.20				& 0.928				& \textbf{5.56} \\ 
		\hline \hline
	\end{tabular}
	\label{Table:ImgQuality}
 \end{table}

\subsection{Comparisons with previous methods}\label{sec:method_robustness}

\begin{figure}[htbp]
	\centering
	\subfloat[The front view of the screen-shooting test.]{
		\label{fig:a}
		\includegraphics[width = 0.21\textwidth]{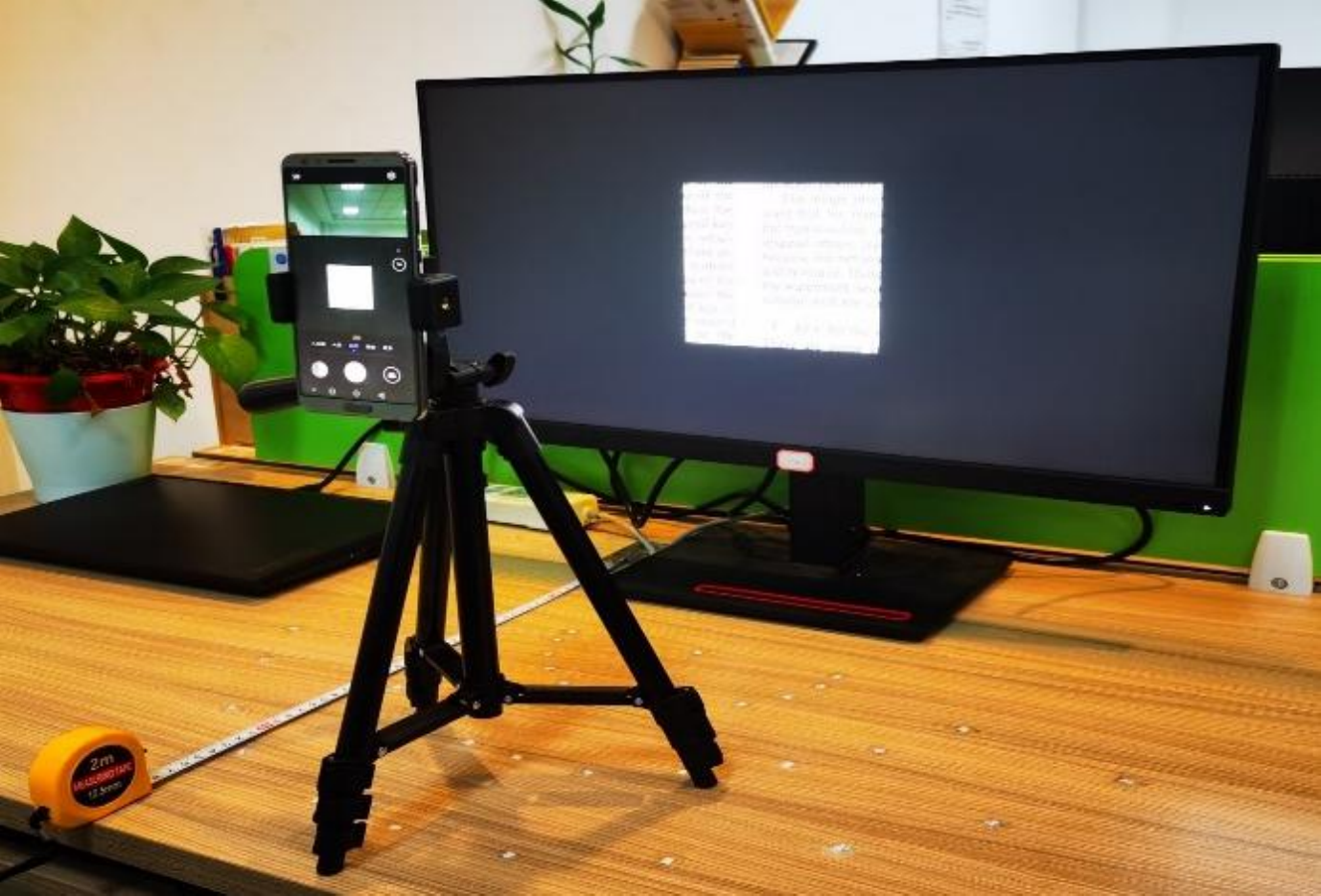}}
	\quad
	\subfloat[The vertical view of the screen-shooting test.]{
		\label{fig:b}
		\includegraphics[width = 0.21\textwidth]{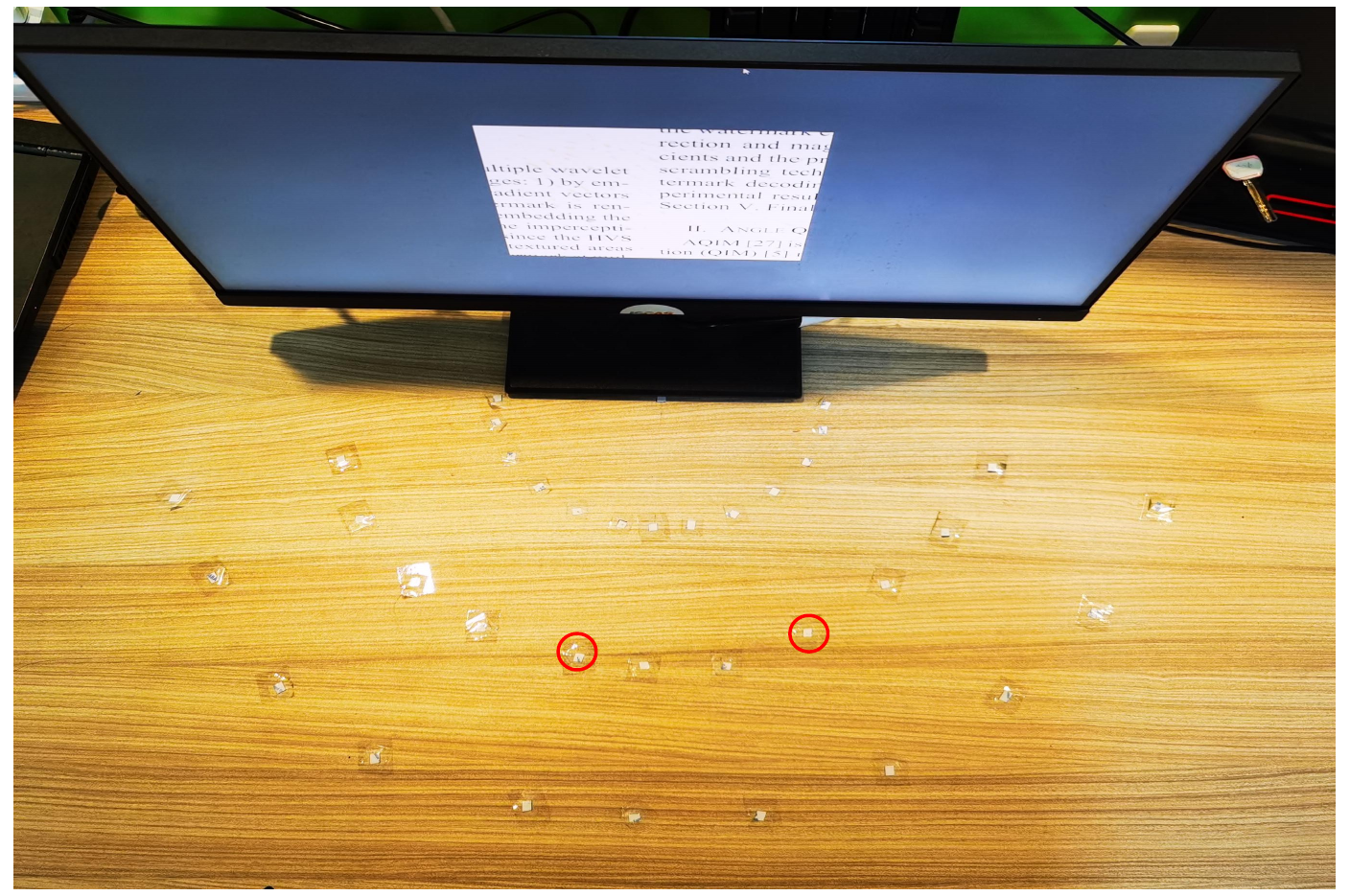}}	
	\caption{
		Experimental scenario of our scheme. As shown in the Fig.~\ref{fig:a}, the environmental scenario is set as similar as possible to the description of \cite{fang2019camera}. As shown in the Fig.~\ref{fig:b}, the dots in the red circle are examples of the shooting point in the experimental test.}
	\label{fig:experimental_scenario}
\end{figure}

 In this section, we will compare the visual quality and robustness with different schemes. Due to the lack of SSR image watermarking algorithms for document, we have to choose three schemes for natural images as comparison schemes. We compare the proposed scheme with the schemes in \cite{fang2018screen} (Transform invariant domain-based), \cite{tancik2020stegastamp} (DNN-based) and \cite{zhang2020udh} (DNN-based). As shown in the Fig.~\ref{fig:experimental_scenario}, the environmental scenario is set as similar as possible to the description of \cite{fang2019camera}. Because the watermark capacity in \cite{fang2019camera} is different from other schemes, so to measure the robustness, we change bit error rate (BER) to average bit accuracy for fair comparison.

\begin{figure}[htbp]
	\includegraphics[width=\linewidth]{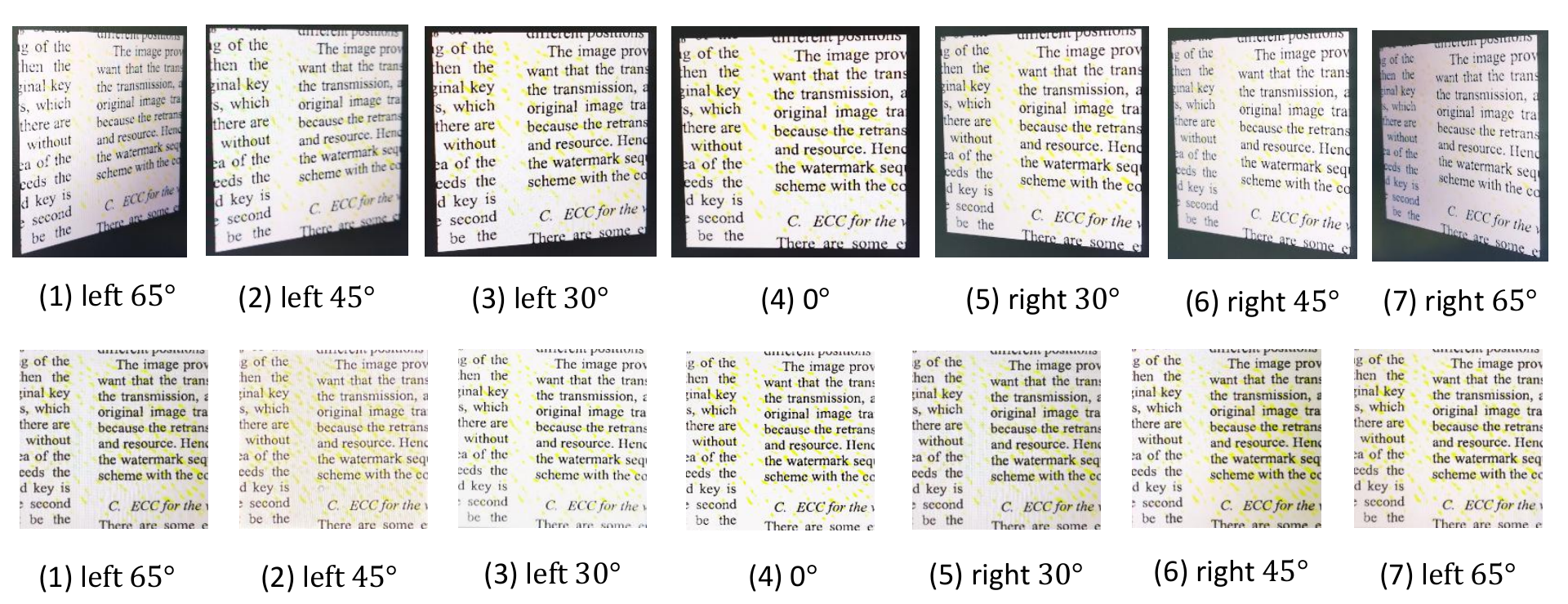}
	\caption{
		Example pairs of the captured images with different horizontal perspective angles (first row) and the corresponding recovered images (second row). These photographs are captured with a Huawei Mate 30 Pro.}
	\label{fig:shooting_angles}
 \end{figure}

\subsubsection{The visual quality with different schemes}\label{sec:visual_quality}

\begin{figure}[htbp]
	\includegraphics[width=\linewidth]{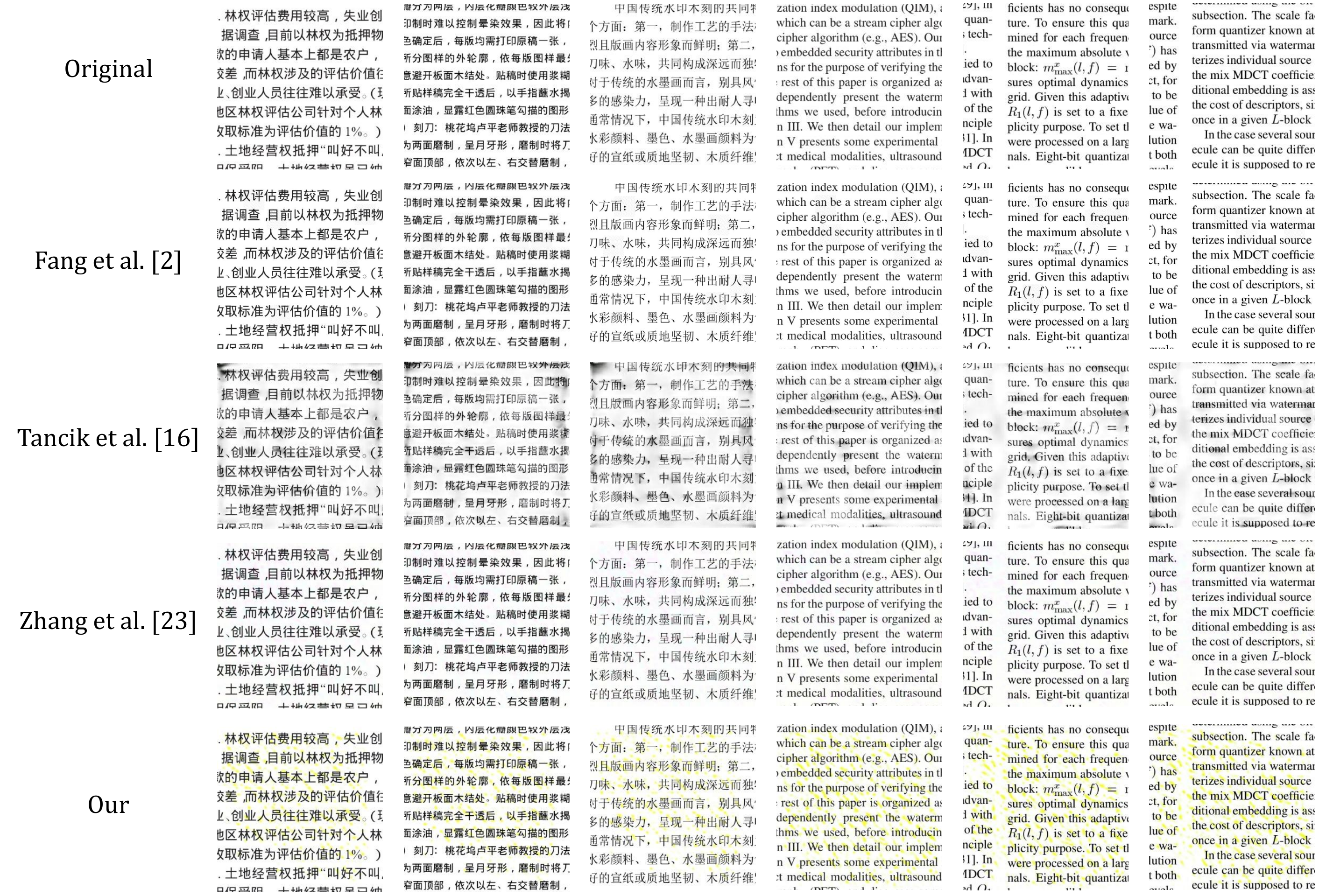}
	\caption{
		Examples of the watermarked images from three state-of-the-arts and our scheme.}
	\label{fig:QualityComparison}
 \end{figure}

 To better evaluate the visual quality, in addition to calculating PSNR and SSIM values, we also perform the \textbf{C}hange Intensity \textbf{P}er Text-\textbf{P}ixel (CPP) test. As is listed in Table~\ref{Table:QualityComparison}, the CPP value of the proposed scheme is much lower than that of other schemes. It indicates that the less modification on the characters of the text pixels on DocImgEN and DocImgCN testsets. In addition, the PSNR values of our scheme are higher than that of two DNN-based methods, which means our scheme has better image quality than two DNN-based methods. Finally, the watermarked images by \cite{fang2018screen}, \cite{tancik2020stegastamp} and \cite{zhang2020udh} are illustrated in Fig~\ref{fig:QualityComparison}.

 \begin{table}[htbp]
	\caption{
		Visual quality of watermarked images with different schemes on DocImgEN and DocImgCN testsets.}
	\centering
	\begin{tabular}{l|ccc}
	\hline \hline
	\makecell[c]{Schemes}											& PSNR (dB) 		& SSIM 				& CPP  \\ \hline
	Fang \emph{et al.} \cite{fang2018screen} (On DocImgEN)			& \textbf{35.42}	& \textbf{0.975}	& 15.12 \\
	Tancik \emph{et al.} \cite{tancik2020stegastamp} (On DocImgEN)	& 22.02				& 0.940				& 81.58 \\
	Zhang \emph{et al.} \cite{zhang2020udh} (On DocImgEN)			& 27.30				& 0.947				& 35.34 \\
	Proposed (On DocImgEN)											& 34.10				& 0.914				& \textbf{6.88} \\ \hline
	Fang \emph{et al.} \cite{fang2018screen} (On DocImgCN)			& \textbf{35.50}	& \textbf{0.977}	& 14.57 \\
	Tancik \emph{et al.} \cite{tancik2020stegastamp} (On DocImgCN)	& 20.30				& 0.922				& 88.58 \\
	Zhang \emph{et al.} \cite{zhang2020udh} (On DocImgCN)			& 28.75				& 0.944				& 27.69 \\
	Proposed (On DocImgCN)											& 33.20				& 0.928				& \textbf{5.56} \\
	\hline \hline
	\end{tabular}
	\label{Table:QualityComparison}
 \end{table}

\subsubsection{The robustness with different schemes}

 In the real scenes, different screen-shooting settings can be used. Therefore, we evaluate the robustness of our scheme under different conditions on DocImgEN and DocImgCN testsets, including different shooting distances, angles (horizontal and vertical perspective angles) and devices.

\begin{figure*}[htbp]
	\includegraphics[width=\linewidth]{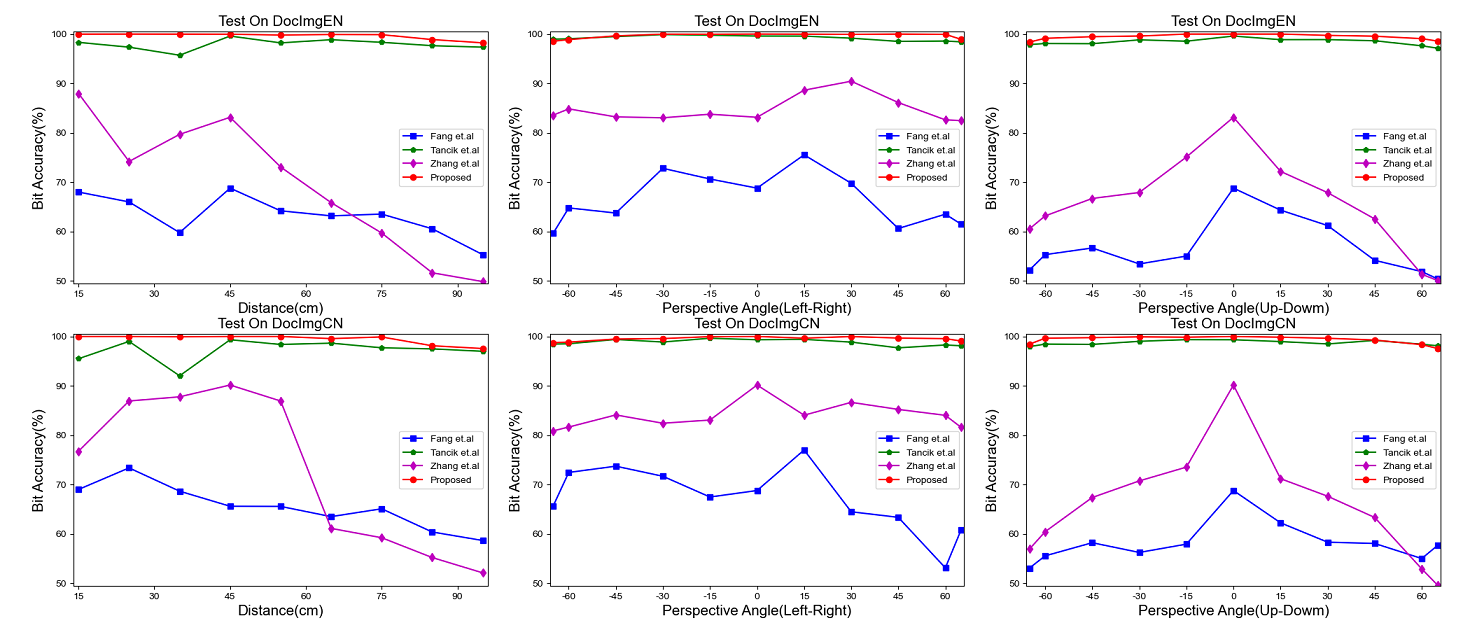}
	\caption{
		Average bit accuracy of the extracted watermarks with different shooting conditions on DocImgEN and DocImgCN testsets.}
	\label{fig:diff_shooting_condition_comparison}
 \end{figure*}

 \textit{1) The Impact of Distance on Robustness:} The average bit accuracy obtained under different shooting distances is listed in Table~\ref{Table:RobustWithDiffDistance}. Compared with natural images, the simple texture of the document would result in the weak strength of I-SIFT key point, so the domain-based method (\cite{fang2018screen}) can not achieve satisfying results as shown in Fig.~\ref{fig:diff_shooting_condition_comparison}. Besides, it is easy to see that our scheme has better performance than other two DNN-based methods in all test distances. When the shooting distance is in a short distance, Moiré pattern occurs in the captured photographs as descripted in \cite{fang2018screen}. However, the Moiré pattern almost has no effect on the robustness of the proposed method, and the bit accuracy is no less than 97\%. Therefore, the proposed scheme has certain robustness to distance change and Moiré pattern.

 \begin{table*}[htbp]
	\caption{
		Average bit accuracy of the extracted watermark with different shooting distances on DocImgEN and DocImgCN testsets.}
	\centering
	\begin{tabular}{l|ccccccccc}
		\hline \hline
		\makecell[c]{Distance (cm)}										 & 15		& 25		& 35		& 45		& 55		& 65		& 75		& 85		& 95 \\ \hline
		Fang \emph{et al.} \cite{fang2018screen} (On DocImgEN)			 & 68.01	& 66.01 	& 59.80 	& 68.80 	& 64.21 	& 63.19 	& 63.55 	& 60.58 	& 55.34 \\
		Tancik \emph{et al.} \cite{tancik2020stegastamp} (On DocImgEN)	 & 98.30	& 97.33 	& 95.70 	& 99.58 	& 98.20 	& 98.84 	& 98.31 	& 97.63 	& 97.33 \\
		Zhang \emph{et al.} \cite{zhang2020udh} (On DocImgEN) 			 & 87.96	& 74.20 	& 79.68 	& 83.14 	& 73.08 	& 65.86 	& 59.72 	& 51.67 	& 49.89 \\
		Proposed (On DocImgEN)			 								 & 99.97	& 99.98 	& 99.97 	& 99.95 	& 99.78 	& 99.89 	& 99.86 	& 98.87 	& 98.21 \\ \hline
		Fang \emph{et al.} \cite{fang2018screen} (On DocImgCN)			 & 69.02	& 73.37 	& 68.67 	& 65.62 	& 65.59 	& 63.52 	& 65.11		& 60.41 	& 58.69 \\
		Tancik \emph{et al.} \cite{tancik2020stegastamp} (On DocImgCN)	 & 95.48	& 98.98 	& 92.02 	& 99.32 	& 98.36 	& 98.63 	& 97.69 	& 97.47 	& 96.98 \\
		Zhang \emph{et al.} \cite{zhang2020udh} (On DocImgCN) 			 & 76.77	& 83.92 	& 87.77 	& 90.15 	& 83.92 	& 61.14 	& 59.25 	& 55.24 	& 52.14 \\
		Proposed (On DocImgCN)			 								 & 99.98	& 99.98 	& 99.92 	& 99.98 	& 99.98 	& 99.55 	& 98.88 	& 98.10 	& 97.54 \\
		\hline \hline
	\end{tabular}
	\label{Table:RobustWithDiffDistance}
 \end{table*}

 \textit{2) The Impact of Horizontal Perspective Angle on Robustness:} Table~\ref{Table:RobustWithDiffHorizontalAngle} shows the bit accuracy of different schemes at the same shooting distance of 40 cm but different horizontal shooting angles. We can see that the robustness of the proposed algorithm is higher than that of other schemes from Left 65$\degree$ to Right 65$\degree$ and the bit accuracy is no less than 98\%. Therefore, the proposed scheme has certain robustness to horizontal perspective angle change.

 \begin{table*}[htbp]
	\scriptsize
	\caption{
		Average bit accuracy of the extracted watermarks with different horizontal perspective angles on DocImgEN and DocImgCN testsets.}
	\centering
	\begin{tabular}{l|ccccccccccc}
		\hline \hline
		\makecell[c]{Horizontal angle ($\degree$)}						& Left 65	& Left 60	& Left 45	& Left 30	& Left 15	& 0 		& Right 15	& Right 30	& Right 45	& Right 60 	& Right 65	\\ \hline
		Fang \emph{et al.} \cite{fang2018screen} (On DocImgEN)			& 59.63 	& 64.00 	& 63.76 	& 72.81 	& 70.05 	& 68.80 	& 75.54 	& 69.79 	& 60.63 	& 63.53 	& 61.52 \\
		Tancik \emph{et al.} \cite{tancik2020stegastamp} (On DocImgEN)	& 98.95 	& 99.03 	& 99.47 	& 99.87		& 99.75		& 99.58 	& 99.57 	& 99.15 	& 98.50		& 98.57		& 98.40 \\
		Zhang \emph{et al.} \cite{zhang2020udh} (On DocImgEN)			& 83.56 	& 84.84 	& 83.22 	& 83.04		& 83.76		& 83.14 	& 88.62 	& 90.44 	& 86.12		& 82.62		& 82.46 \\ 
		Proposed (On DocImgEN)											& 98.53 	& 98.86		& 99.62 	& 99.96		& 99.94		& 99.98 	& 99.96 	& 99.92 	& 99.98		& 99.92		& 98.92 \\ \hline
		Fang \emph{et al.} \cite{fang2018screen} (On DocImgCN)			& 65.62 	& 72.46 	& 73.73 	& 71.69 	& 67.49 	& 68.80 	& 77.01 	& 64.50 	& 63.38 	& 53.15 	& 60.81 \\
		Tancik \emph{et al.} \cite{tancik2020stegastamp} (On DocImgCN)	& 98.44 	& 98.50 	& 99.38 	& 98.86 	& 99.62 	& 99.32 	& 99.40 	& 98.84 	& 97.68		& 98.26		& 98.10 \\
		Zhang \emph{et al.} \cite{zhang2020udh} (On DocImgCN)			& 80.87 	& 81.65 	& 84.10 	& 82.42 	& 83.07 	& 90.15 	& 84.05 	& 86.65 	& 85.22		& 84.02		& 81.60 \\ 
		Proposed (On DocImgCN)											& 98.72 	& 98.84 	& 99.46 	& 99.56 	& 99.94 	& 99.98 	& 99.64 	& 99.96 	& 99.66		& 99.54		& 99.06 \\ 
		\hline \hline
	\end{tabular}
	\label{Table:RobustWithDiffHorizontalAngle}
 \end{table*}

 \textit{3) The Impact of Vertical Perspective Angle on Robustness:} Table~\ref{Table:RobustWithDiffVerticalAngle} show that the bit accuracy obtained by different schemes at the same shooting distance of 40 cm but different vertical shooting angles. It is easy to see that the robustness of the proposed algorithm is higher than that of other schemes from Up 65$\degree$ to Down 65$\degree$ and the bit accuracy is no less than 97\%. Therefore, the proposed scheme has certain robustness to vertical perspective angle change.

 \begin{table*}[htbp]
	\scriptsize
	\caption{
		Average bit accuracy of the extracted watermarks with different vertical perspective angles on DocImgEN and DocImgCN testsets.}
	\centering
	\begin{tabular}{l|ccccccccccc}
		\hline \hline
		\makecell[c]{Vertical angle ($\degree$)}						& Up 65 	& Up 60 	& Up 45 	& Up 30 	& Up 15 	& 0 		& Down 15 	& Down 30	& Down 45 	& Down 60 	& Down 65\\ \hline
		Fang \emph{et al.} \cite{fang2018screen} (On DocImgEN)			& 52.23 	& 55.34 	& 56.69 	& 53.46 	& 55.05 	& 68.80 	& 64.35		& 61.23		& 54.19 	& 51.91 	& 50.45 \\
		Tancik \emph{et al.} \cite{tancik2020stegastamp} (On DocImgEN)	& 97.86 	& 98.07 	& 98.03 	& 98.81 	& 98.54 	& 99.58 	& 98.85		& 98.86 	& 98.63 	& 97.60 	& 97.13 \\
		Zhang \emph{et al.} \cite{zhang2020udh} (On DocImgEN)			& 60.56 	& 63.18 	& 66.71 	& 67.92 	& 75.09 	& 83.14 	& 72.29		& 67.89 	& 62.57 	& 51.37 	& 50.12 \\ 
		Proposed (On DocImgEN)											& 98.37 	& 99.12		& 99.45 	& 99.58 	& 99.98 	& 99.98 	& 99.98 	& 99.72 	& 99.55 	& 99.05 	& 98.57 \\ \hline
		Fang \emph{et al.} \cite{fang2018screen} (On DocImgCN)			& 53.11 	& 55.58 	& 58.24 	& 56.27 	& 57.95 	& 68.80 	& 62.27 	& 58.35		& 58.09 	& 55.04 	& 57.67 \\
		Tancik \emph{et al.} \cite{tancik2020stegastamp} (On DocImgCN)	& 97.93 	& 98.43 	& 98.37 	& 99.00 	& 99.34 	& 99.32 	& 98.95 	& 98.47 	& 99.18 	& 98.41 	& 98.10 \\
		Zhang \emph{et al.} \cite{zhang2020udh} (On DocImgCN)			& 56.95 	& 60.47 	& 67.39 	& 70.75 	& 73.54 	& 90.15 	& 71.20 	& 67.65 	& 63.35 	& 52.94 	& 49.68 \\ 
		Proposed (On DocImgCN)											& 98.42 	& 99.63 	& 99.74 	& 99.90 	& 99.83 	& 99.98 	& 99.83 	& 99.62 	& 99.26 	& 98.33 	& 97.56 \\
		\hline \hline
	\end{tabular}
	\label{Table:RobustWithDiffVerticalAngle}
 \end{table*}

 \textit{4) The Impact of Different Devices on Robustness:} To test the robustness for different devices, we use multiple device combinations (phone-display) in the testing stage. To display watermarked images, we choose five types of displays, namely AOC Q271PQ, HP P224, AOC Q2490PXQ, ThinkVision T27h-20, and Dell S2721DS. To take photographs, we choose five phone devices with cameras, namely Huawei nova 2s, Huawei mate 30 pro, MEIZU M3X, Mi 8 UD, and IPhone 11 Pro Max. For each combination of display-phone tests, the shooting angle is 0 degrees and shooting distance is 40 cm. The bit accuracy obtained by different combinations of devices is listed in Table~\ref{Table:RobustWithDiffDeviceOnDocImgENAndDocImgCN}. For all combination of display-phone tests, the extraction accuracy is no less than 97\%, it indicates the scheme has good applicability to phones and displays used in the testing stage.

 \begin{table*}[htbp]
	\centering
	\caption{
		Average bit accuracy of the extracted watermarks with different devices on DocImgEN testset.}
	\begin{tabular}{l|ccccc}
	\hline \hline
	\makecell[c]{\diagbox{Screens}{Phones}}	& Huawei Nova 2S	& Huawei Mate 30 Pro	& MEIZU M3X		& Mi 8 UD		& IPhone 11 Pro Max \\ \hline
	HP P224 (On DocImgEN)					& 99.55 			& 99.93 				& 99.16 		& 98.90 		& 98.41 \\
	CHANGHONG 27P620F (On DocImgEN)			& 99.69 			& 99.84 				& 98.91 		& 98.64 		& 98.88 \\
	AOC Q271PQ (On DocImgEN)				& 97.46 			& 99.99 				& 99.81 		& 99.95 		& 99.91 \\
	Dell U2720Q (On DocImgEN)				& 99.51 			& 99.95 				& 98.05 		& 98.04 		& 99.99 \\
	ThinkVision T27h-20 (On DocImgEN)		& 99.86 			& 99.98 				& 99.83 		& 99.68 		& 99.33 \\ \hline
	HP P224 (On DocImgCN)					& 98.96 			& 99.62 				& 98.59 		& 97.63 		& 98.66 \\
	CHANGHONG 27P620F (On DocImgCN)			& 99.02 			& 99.15 				& 98.46 		& 98.43 		& 99.24 \\
	AOC Q271PQ (On DocImgCN)				& 99.64 			& 99.91 				& 98.59 		& 99.19 		& 98.10 \\
	Dell U2720Q (On DocImgCN)				& 99.03 			& 99.42 				& 97.38 		& 98.23 		& 99.89 \\
	ThinkVision T27h-20 (On DocImgCN)		& 99.24 			& 99.98 				& 98.65 		& 99.66 		& 99.99 \\
	\hline \hline
	\end{tabular}
    \label{Table:RobustWithDiffDeviceOnDocImgENAndDocImgCN}
 \end{table*}

\subsection{Further improvement by adjusting embedding strength}\label{sec:improvement}

 Although the PSNR and SSIM values in Table~\ref{Table:QualityComparison} is acceptable, the examples in Fig.~\ref{fig:QualityComparison} still shows clear traces due to the obvious contrast between the foreground and background in the document image. So, we try to alleviate this phenomenon by adjusting the embedding strength in the training and testing process.
 
 According to the Formula~\ref{equ:addition}, the adjustment of the embedding strength factors $\alpha$ can influence the quality of watermarked image and the bit accuracy. As shown in Table~\ref{Table:RobustWithDiffHorizontalAngle} and Table~\ref{Table:RobustWithDiffVerticalAngle}, our scheme achieves high bit accuracy during testing in real scenes. It means that redundancy space is existed during training, the quality of watermark image can be improved by decreasing the embedding strength factors $\alpha$ without much loss in the bit accuracy. It indicates that we can set a higher embedding strength during training, and decrease the embedding strength to apply in the real scenes after the model is fully trained.
 
\begin{figure}[htbp]
	\includegraphics[width=\linewidth]{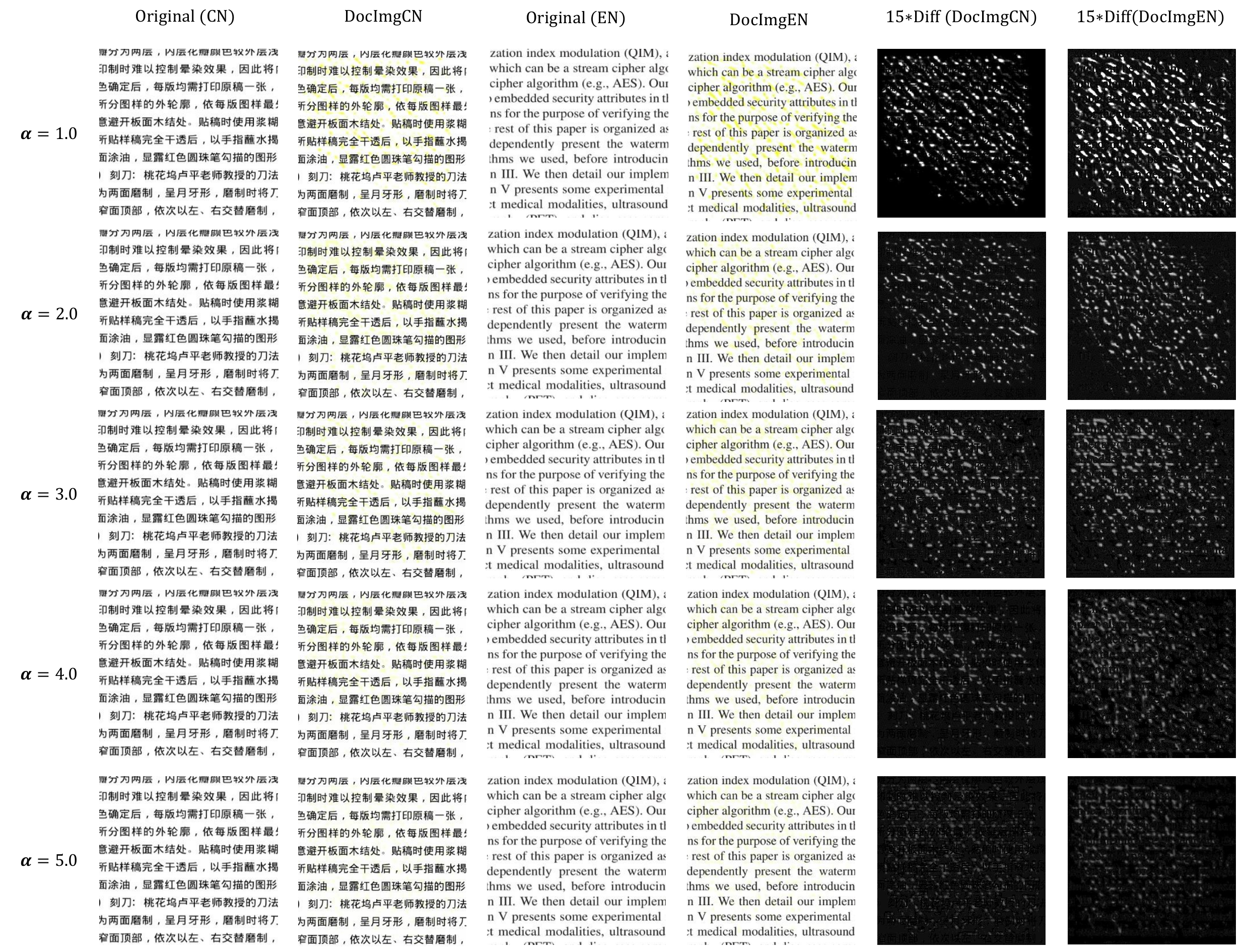}
	\caption{
		Visual examples of our scheme with different embedding strength factors during the testing stage on DocImgEN and DocImgCN testsets.}
	\label{fig:QualityWithDifEmbeddingFactor}
 \end{figure}

 In this subsection, we try to set the embedding strength factors $\alpha=1.0, 2.0, 3.0, 4.0, 5.0$ in the training process and keep $\alpha$ to be 1.0 in the testing process. As shown in Table~\ref{Table:QualityWithDiffEmbeddingFactor} and Fig.~\ref{fig:QualityWithDifEmbeddingFactor}, with larger initial $\alpha$, better image quality is achieved. While as shown in Table~\ref{Table:RobustWithDiffDistanceAndEmbeddingFactor}, Table~\ref{Table:RobustWithDiffHorizontalAngleAndEmbeddingFactor} and Table~\ref{Table:RobustWithDiffVerticalAngleAndEmbeddingFactor}, the bit accuracy is not decreased a lot, including in different shooting distances, horizontal perspective angles, and vertical perspective angles tests.

 \begin{table}[htbp]
	\caption{
		PSNR, SSIM and CPP of our scheme with different embedding strength factors during the testing stage on DocImgEN and DocImgCN testsets.}
	\centering
	\begin{tabular}{l|ccc}
		\hline \hline
		\makecell[c]{Schemes}			& PSNR (dB)		& SSIM		& CPP \\ \hline
		On DocImgEN with $\alpha$=1.0	& 34.10			& 0.914		& 6.88 \\
		On DocImgEN with $\alpha$=2.0	& 38.50			& 0.938		& 4.89 \\
		On DocImgEN with $\alpha$=3.0	& 39.40			& 0.973		& 3.50 \\
		On DocImgEN with $\alpha$=4.0	& 42.10			& 0.949		& 3.32 \\
		On DocImgEN with $\alpha$=5.0	& 43.70			& 0.963		& 2.95 \\ \hline
		On DocImgCN with $\alpha$=1.0	& 33.20			& 0.928		& 5.56 \\
		On DocImgCN with $\alpha$=2.0	& 38.90			& 0.937		& 5.13 \\
		On DocImgCN with $\alpha$=3.0	& 39.50			& 0.950		& 4.34 \\
		On DocImgCN with $\alpha$=4.0	& 40.90			& 0.957		& 3.42 \\
		On DocImgCN with $\alpha$=5.0	& 42.58			& 0.966		& 3.51 \\
		\hline \hline
	\end{tabular}
	\label{Table:QualityWithDiffEmbeddingFactor}
 \end{table}

 \begin{table*}[htbp]
	\caption{
		Average bit accuracy of our scheme with the different embedding strength factors and distances during the testing process on DocImgEN and DocImgCN testsets.}
	\centering
	\begin{tabular}{l|ccccccccc}
		\hline \hline
		\makecell[c]{Distance (cm)}		& 15 	& 25 	& 35 	& 45 	& 55 	& 65 	& 75	& 85	& 95  \\ \hline
		On DocImgEN with $\alpha$=1.0	& 99.97	& 99.98	& 99.97	& 99.95	& 99.78	& 99.89	& 99.86	& 98.87	& 97.33 \\
		On DocImgEN with $\alpha$=2.0 	& 99.92	& 99.90	& 98.35	& 99.93	& 99.80	& 99.56	& 99.39	& 98.59	& 97.02 \\
		On DocImgEN with $\alpha$=3.0 	& 99.22	& 99.36	& 97.48	& 99.48	& 99.28	& 98.13	& 96.74	& 96.84	& 95.25 \\ 
		On DocImgEN with $\alpha$=4.0 	& 98.76	& 98.92	& 98.96	& 98.53	& 98.96	& 98.64	& 97.66	& 96.83	& 96.10 \\
		On DocImgEN with $\alpha$=5.0	& 95.16	& 98.50	& 97.86	& 97.60	& 97.74	& 97.51	& 96.63	& 96.29	& 94.32 \\ \hline
		On DocImgCN with $\alpha$=1.0	& 99.98	& 99.98	& 99.92	& 99.98	& 99.98	& 99.55	& 98.88	& 98.10	& 96.98 \\
		On DocImgCN with $\alpha$=2.0 	& 99.74	& 99.45	& 98.60	& 99.53	& 98.70	& 98.64	& 98.43	& 98.48	& 97.60 \\
		On DocImgCN with $\alpha$=3.0 	& 99.70	& 98.85	& 97.30	& 98.90	& 98.70	& 98.58	& 98.40	& 98.21	& 96.01 \\ 
		On DocImgCN with $\alpha$=4.0 	& 97.35	& 97.50	& 94.26	& 97.83	& 98.26	& 97.91	& 97.78	& 95.83	& 95.01 \\
		On DocImgCN with $\alpha$=5.0	& 93.22	& 95.36	& 94.92	& 96.59	& 95.02	& 94.37	& 95.77	& 94.68	& 94.82 \\
		\hline \hline
	\end{tabular}
	\label{Table:RobustWithDiffDistanceAndEmbeddingFactor}
 \end{table*}

 \begin{table*}[htbp]
	\scriptsize
	\caption{
		Average bit accuracy of our scheme with the different embedding strength factors and horizontal perspective angles during the testing stage on DocImgEN and DocImgCN testsets.}
	\centering
	\begin{tabular}{l|ccccccccccc}
		\hline \hline
		\makecell[c]{Horizontal angle ($\degree$)}	& Left 65	& Left 60	& Left 45	& Left 30	& Left 15	& 0 		& Right 15	& Right 30	& Right 45	& Right 60	& Right 65 \\ \hline
		On DocImgEN with $\alpha$=1.0				& 98.53		& 98.86		& 99.62		& 99.96		& 99.94		& 99.98		& 99.96		& 99.92		& 99.98		& 99.92		& 98.92 \\
		On DocImgEN with $\alpha$=2.0 				& 97.45		& 98.95		& 99.10		& 99.85		& 99.72		& 99.93		& 99.33		& 99.42		& 99.27		& 99.10		& 97.62 \\
		On DocImgEN with $\alpha$=3.0 				& 97.16		& 98.62		& 99.18 	& 99.54		& 99.50		& 99.48		& 98.54		& 99.08		& 99.02		& 98.52		& 97.06 \\ 
		On DocImgEN with $\alpha$=4.0 				& 96.91		& 98.96		& 98.96		& 98.13		& 98.29		& 98.53		& 98.48		& 98.36		& 98.29		& 98.54		& 96.60 \\
		On DocImgEN with $\alpha$=5.0				& 96.02		& 97.30		& 98.86		& 98.82		& 98.76		& 97.60		& 98.68		& 98.86		& 98.92		& 96.76		& 95.62 \\ \hline
		On DocImgCN with $\alpha$=1.0				& 98.72		& 98.84		& 99.46		& 99.56		& 99.94		& 99.98 	& 99.64		& 99.96		& 99.66		& 99.54		& 99.06 \\
		On DocImgCN with $\alpha$=2.0 				& 98.35		& 98.69		& 99.80		& 99.94		& 99.80		& 99.53		& 99.91		& 99.91		& 99.77		& 99.54		& 98.66 \\
		On DocImgCN with $\alpha$=3.0 				& 97.72		& 97.73		& 99.81		& 99.82		& 99.87		& 98.90		& 99.86		& 99.89		& 99.83		& 98.86		& 96.81 \\ 
		On DocImgCN with $\alpha$=4.0 				& 96.41		& 97.53		& 98.63		& 98.61		& 98.52		& 97.83		& 98.71		& 98.60		& 98.83		& 98.59		& 96.55 \\
		On DocImgCN with $\alpha$=5.0				& 95.38		& 95.19		& 96.60		& 96.15		& 97.47		& 97.59 	& 97.10		& 96.86		& 96.48		& 96.38		& 96.06 \\
		\hline \hline
	\end{tabular}
	\label{Table:RobustWithDiffHorizontalAngleAndEmbeddingFactor}
 \end{table*}
  
 \begin{table*}[htbp]
	\scriptsize
	\caption{
		Average bit accuracy of our scheme with the different embedding strength factors and vertical perspective angles during the testing stage on DocImgEN and DocImgCN testsets. }
	\centering
	\begin{tabular}{l|ccccccccccc}
		\hline \hline
		\makecell[c]{Vertical angle ($\degree$)}	& Up 65 	& Up 60 	& Up 45 	& Up 30 	& Up 15 	& 0 		& Down 15	& Down 30	& Down 45	& Down 60	& Down 65\\ \hline
		On DocImgEN with $\alpha$=1.0				& 98.37 	& 99.12		& 99.45 	& 99.58 	& 99.98 	& 99.98 	& 99.98 	& 99.72 	& 99.55 	& 99.05 	& 98.57 \\
		On DocImgEN with $\alpha$=2.0 				& 97.02 	& 97.50 	& 98.58 	& 98.89 	& 99.86 	& 99.93 	& 99.58 	& 98.84 	& 98.87 	& 98.32 	& 97.85 \\
		On DocImgEN with $\alpha$=3.0 				& 96.14 	& 96.51 	& 97.64 	& 98.70 	& 98.88 	& 99.48 	& 98.52 	& 97.73		& 96.12 	& 95.52 	& 94.33 \\ 
		On DocImgEN with $\alpha$=4.0 				& 96.33 	& 96.92 	& 97.35 	& 98.86 	& 98.36 	& 98.53 	& 97.97 	& 98.33 	& 95.89 	& 94.36 	& 94.10 \\
		On DocImgEN with $\alpha$=5.0				& 93.15 	& 94.20 	& 96.76 	& 97.45 	& 97.69 	& 97.60 	& 96.57 	& 95.88 	& 96.51 	& 93.11 	& 92.82 \\ \hline
		On DocImgCN with $\alpha$=1.0				& 98.42 	& 99.63 	& 99.74 	& 99.90 	& 99.83 	& 99.98 	& 99.83 	& 99.62 	& 99.26 	& 98.33 	& 97.56 \\
		On DocImgCN with $\alpha$=2.0 				& 98.17 	& 98.90 	& 98.97 	& 99.54 	& 99.94 	& 99.53 	& 99.80 	& 99.50 	& 98.93 	& 97.70 	& 97.17 \\
		On DocImgCN with $\alpha$=3.0 				& 97.51 	& 97.56 	& 98.60 	& 98.51 	& 98.52 	& 98.90 	& 98.64 	& 98.61 	& 97.41 	& 97.96 	& 97.02 \\ 
		On DocImgCN with $\alpha$=4.0 				& 96.55 	& 96.90 	& 97.10 	& 97.47 	& 98.57 	& 97.83 	& 97.41 	& 97.76 	& 96.57 	& 96.03 	& 95.83 \\
		On DocImgCN with $\alpha$=5.0				& 95.39 	& 96.54 	& 97.71 	& 96.60 	& 97.08 	& 97.59 	& 96.87 	& 97.44 	& 96.32 	& 94.03 	& 94.96 \\
		\hline \hline
	\end{tabular}
	\label{Table:RobustWithDiffVerticalAngleAndEmbeddingFactor}
 \end{table*}

\section{Conclusions}\label{sec:conclusion}
 In this paper, we propose an end-to-end screen-shooting resilient document image watermarking scheme using deep neural network. Our scheme consists of three core components: Encoder, Distortion Layer, and Decoder. The Encoder is designed to embed watermark into document images. The Distortion Layer is added to compensate a series of distortions in the screen-shooting process (such as camera distortion, shooting distortion, and light source distortion), which enables the watermarked image can be resilient to the distortions in real screen-shooting scenes. The Decoder is designed to extract the watermark from the captured photographs. Besides, An embedding strength adjustment strategy is discusses to decrease the embedding trace with little loss of extraction accuracy, and the experimental results show the feasibility of our proposed scheme.

\section*{Acknowledgment}\label{sec:acknowledgment}
This work is supported in part by the National Key Research and Development Plan of China under Grant 2020YFB1005600, in part by the National Natural Science Foundation of China under grant numbers 62122032, 62172233, 62102189, U1936118, 61825203, U1736203, and 61732021, in part by the Major Program of Guangdong Basic and Applied Research Project under Grant 2019B030302008, in part by Six Peak Talent project of Jiangsu Province (R2016L13), Qinglan Project of Jiangsu Province, and `333` project of Jiangsu Province, in part by the National Joint Engineering Research Center for Network Security Detection and Protection Technology, in part by the Priority Academic Program Development of Jiangsu Higher Education Institutions (PAPD) fund, in part by the Collaborative Innovation Center of Atmospheric Environment and Equipment Technology (CICAEET) fund, China. Zhihua Xia is supported by BK21+ program from the Ministry of Education of Korea.

\ifCLASSOPTIONcaptionsoff
	\newpage
\fi
\bibliographystyle{unsrt}
\bibliography{SSRDIW}

\begin{thebibliography}{10}

\bibitem{gugelmann2018screen}
David Gugelmann, David Sommer, Vincent Lenders, Markus Happe, and Laurent
  Vanbever.
\newblock Screen watermarking for data theft investigation and attribution.
\newblock In {\em 2018 10th International Conference on Cyber Conflict
  (CyCon)}, pages 391--408. IEEE, 2018.

\bibitem{fang2018screen}
Han Fang, Weiming Zhang, Hang Zhou, Hao Cui, and Nenghai Yu.
\newblock Screen-shooting resilient watermarking.
\newblock {\em IEEE Transactions on Information Forensics and Security},
  14(6):1403--1418, 2018.

\bibitem{pereira2000robust}
Shelby Pereira and Thierry Pun.
\newblock Robust template matching for affine resistant image watermarks.
\newblock {\em IEEE transactions on image Processing}, 9(6):1123--1129, 2000.

\bibitem{kang2003dwt}
Xiangui Kang, Jiwu Huang, Yun~Q Shi, and Yan Lin.
\newblock A dwt-dft composite watermarking scheme robust to both affine
  transform and jpeg compression.
\newblock {\em IEEE transactions on circuits and systems for video technology},
  13(8):776--786, 2003.

\bibitem{pramila2007multiple}
Anu Pramila, Anja Keskinarkaus, and Tapio Sepp{\"a}nen.
\newblock Multiple domain watermarking for print-scan and jpeg resilient data
  hiding.
\newblock In {\em International Workshop on Digital Watermarking}, pages
  279--293. Springer, 2007.

\bibitem{he2005practical}
Dajun He and Qibin Sun.
\newblock A practical print-scan resilient watermarking scheme.
\newblock In {\em IEEE International Conference on Image Processing 2005},
  volume~1, pages I--257. IEEE, 2005.

\bibitem{solanki2006print}
Kaushal Solanki, Upamanyu Madhow, BS~Manjunath, Shiv Chandrasekaran, and
  Ibrahim El-Khalil.
\newblock Print and scan'resilient data hiding in images.
\newblock {\em IEEE Transactions on Information Forensics and Security},
  1(4):464--478, 2006.

\bibitem{kang2010efficient}
Xiangui Kang, Jiwu Huang, and Wenjun Zeng.
\newblock Efficient general print-scanning resilient data hiding based on
  uniform log-polar mapping.
\newblock {\em IEEE Transactions on Information Forensics and Security},
  5(1):1--12, 2010.

\bibitem{amiri2014robust}
S~Hamid Amiri and Mansour Jamzad.
\newblock Robust watermarking against print and scan attack through efficient
  modeling algorithm.
\newblock {\em Signal Processing: Image Communication}, 29(10):1181--1196,
  2014.

\bibitem{nakamura2004fast}
Takao Nakamura, Atsushi Katayama, Masashi Yamamuro, and Noboru Sonehara.
\newblock Fast watermark detection scheme for camera-equipped cellular phone.
\newblock In {\em Proceedings of the 3rd international conference on Mobile and
  ubiquitous multimedia}, pages 101--108, 2004.

\bibitem{kim2006image}
Won-gyum Kim, Seon~Hwa Lee, and Yong-seok Seo.
\newblock Image fingerprinting scheme for print-and-capture model.
\newblock In {\em Pacific-Rim Conference on Multimedia}, pages 106--113.
  Springer, 2006.

\bibitem{pramila2012toward}
Anu Pramila, Anja Keskinarkaus, and Tapio Sepp{\"a}nen.
\newblock Toward an interactive poster using digital watermarking and a mobile
  phone camera.
\newblock {\em Signal, Image and Video Processing}, 6(2):211--222, 2012.

\bibitem{delgado2013digital}
Lorenzo~Antonio Delgado-Guillen, Jose~Juan Garcia-Hernandez, and Cesar
  Torres-Huitzil.
\newblock Digital watermarking of color images utilizing mobile platforms.
\newblock In {\em 2013 IEEE 56th International Midwest Symposium on Circuits
  and Systems (MWSCAS)}, pages 1363--1366. IEEE, 2013.

\bibitem{gourrame2016robust}
Khadija Gourrame, Hassan Douzi, Rachid Harba, Frederic Ros, Mohamed El~Hajji,
  Rabia Riad, and Meina Amar.
\newblock Robust print-cam image watermarking in fourier domain.
\newblock In {\em International Conference on Image and Signal Processing},
  pages 356--365. Springer, 2016.

\bibitem{liang2019robust}
Shuang Liang and Xingjun Wang.
\newblock Robust image watermarking in the print-cam process.
\newblock In {\em 2019 IEEE 19th International Conference on Communication
  Technology (ICCT)}, pages 1626--1630. IEEE, 2019.

\bibitem{tancik2020stegastamp}
Matthew Tancik, Ben Mildenhall, and Ren Ng.
\newblock Stegastamp: Invisible hyperlinks in physical photographs.
\newblock In {\em Proceedings of the IEEE/CVF Conference on Computer Vision and
  Pattern Recognition}, pages 2117--2126, 2020.

\bibitem{jia2020rihoop}
Jun Jia, Zhongpai Gao, Kang Chen, Menghan Hu, Xiongkuo Min, Guangtao Zhai, and
  Xiaokang Yang.
\newblock Rihoop: Robust invisible hyperlinks in offline and online
  photographs.
\newblock {\em IEEE Transactions on Cybernetics}, 2020.

\bibitem{fridrich2009digital}
Jessica Fridrich.
\newblock Digital image forensics.
\newblock {\em IEEE Signal Processing Magazine}, 26(2):26--37, 2009.

\bibitem{fang2019camera}
Han Fang, Weiming Zhang, Zehua Ma, Hang Zhou, Shan Sun, Hao Cui, and Nenghai
  Yu.
\newblock A camera shooting resilient watermarking scheme for underpainting
  documents.
\newblock {\em IEEE Transactions on Circuits and Systems for Video Technology},
  30(11):4075--4089, 2019.

\bibitem{li2021screen}
Li~Li, Rui Bai, Shanqing Zhang, Chin-Chen Chang, and Mengtao Shi.
\newblock Screen-shooting resilient watermarking scheme via learned invariant
  keypoints and qt.
\newblock {\em Sensors}, 21(19):6554, 2021.

\bibitem{wengrowski2019light}
Eric Wengrowski and Kristin Dana.
\newblock Light field messaging with deep photographic steganography.
\newblock In {\em Proceedings of the IEEE/CVF Conference on Computer Vision and
  Pattern Recognition}, pages 1515--1524, 2019.

\bibitem{zhang2021brief}
Chaoning Zhang, Chenguo Lin, Philipp Benz, Kejiang Chen, Weiming Zhang, and
  In~So Kweon.
\newblock A brief survey on deep learning based data hiding, steganography and
  watermarking.
\newblock {\em arXiv preprint arXiv:2103.01607}, 2021.

\bibitem{zhang2020udh}
Chaoning Zhang, Philipp Benz, Adil Karjauv, Geng Sun, and In~So Kweon.
\newblock Udh: Universal deep hiding for steganography, watermarking, and light
  field messaging.
\newblock {\em Advances in Neural Information Processing Systems},
  33:10223--10234, 2020.

\bibitem{fang2020deep}
Han Fang, Dongdong Chen, Qidong Huang, Jie Zhang, Zehua Ma, Weiming Zhang, and
  Nenghai Yu.
\newblock Deep template-based watermarking.
\newblock {\em IEEE Transactions on Circuits and Systems for Video Technology},
  31(4):1436--1451, 2020.

\bibitem{fang2021tera}
Han Fang, Dongdong Chen, Feng Wang, Zehua Ma, Honggu Liu, Wenbo Zhou, Weiming
  Zhang, and Neng-Hai Yu.
\newblock Tera: Screen-to-camera image code with transparency, efficiency,
  robustness and adaptability.
\newblock {\em IEEE Transactions on Multimedia}, 2021.

\bibitem{vaswani2017attention}
Ashish Vaswani, Noam Shazeer, Niki Parmar, Jakob Uszkoreit, Llion Jones,
  Aidan~N Gomez, {\L}ukasz Kaiser, and Illia Polosukhin.
\newblock Attention is all you need.
\newblock {\em Advances in neural information processing systems}, 30, 2017.

\bibitem{hasinoff2014photon}
Samuel~W. Hasinoff.
\newblock Photon, poisson noise.
\newblock In {\em Computer Vision, A Reference Guide}, 2014.

\bibitem{ronneberger2015u}
Olaf Ronneberger, Philipp Fischer, and Thomas Brox.
\newblock U-net: Convolutional networks for biomedical image segmentation.
\newblock In {\em International Conference on Medical image computing and
  computer-assisted intervention}, pages 234--241. Springer, 2015.

\bibitem{shin2017jpeg}
Richard Shin and Dawn Song.
\newblock Jpeg-resistant adversarial images.
\newblock In {\em NIPS 2017 Workshop on Machine Learning and Computer
  Security}, volume~1, 2017.

\bibitem{zhao2016loss}
Hang Zhao, Orazio Gallo, Iuri Frosio, and Jan Kautz.
\newblock Loss functions for image restoration with neural networks.
\newblock {\em IEEE Transactions on computational imaging}, 3(1):47--57, 2016.

\bibitem{ge2022robust}
Sulong Ge, Zhihua Xia, Jianwei Fei, Xingming Sun, and Jian Weng.
\newblock A robust document image watermarking scheme using deep neural
  network.
\newblock {\em arXiv preprint arXiv:2202.13067}, 2022.

\bibitem{kingma2014adam}
Diederik~P Kingma and Jimmy Ba.
\newblock Adam: A method for stochastic optimization.
\newblock {\em arXiv preprint arXiv:1412.6980}, 2014.

\bibitem{wang2004image}
Zhou Wang, Alan~C Bovik, Hamid~R Sheikh, and Eero~P Simoncelli.
\newblock Image quality assessment: from error visibility to structural
  similarity.
\newblock {\em IEEE transactions on image processing}, 13(4):600--612, 2004.

\end{thebibliography}
\vfill

\begin{IEEEbiography}[{\includegraphics[width=1in,height=1.25in,clip,keepaspectratio]{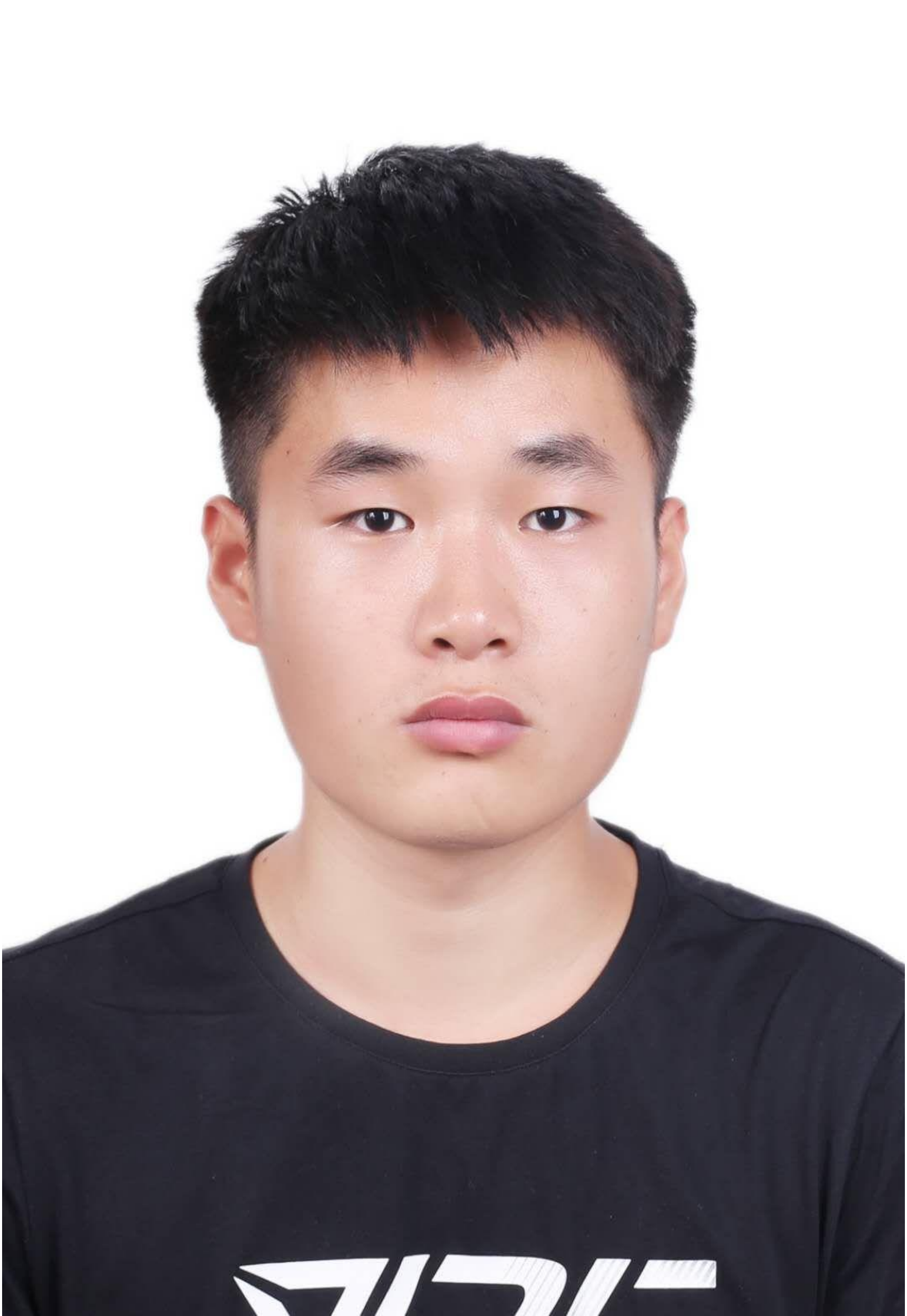}}]{Sulong Ge} received his BE degree in Software Engineering from TianGong University in 2020. He is currently pursuing master degree in School of Computer Science in Nanjing University of Information Science and Technology. His research interests include data hiding and information forensics.
\end{IEEEbiography}

\begin{IEEEbiography}[{\includegraphics[width=1in,height=1.25in,clip,keepaspectratio]{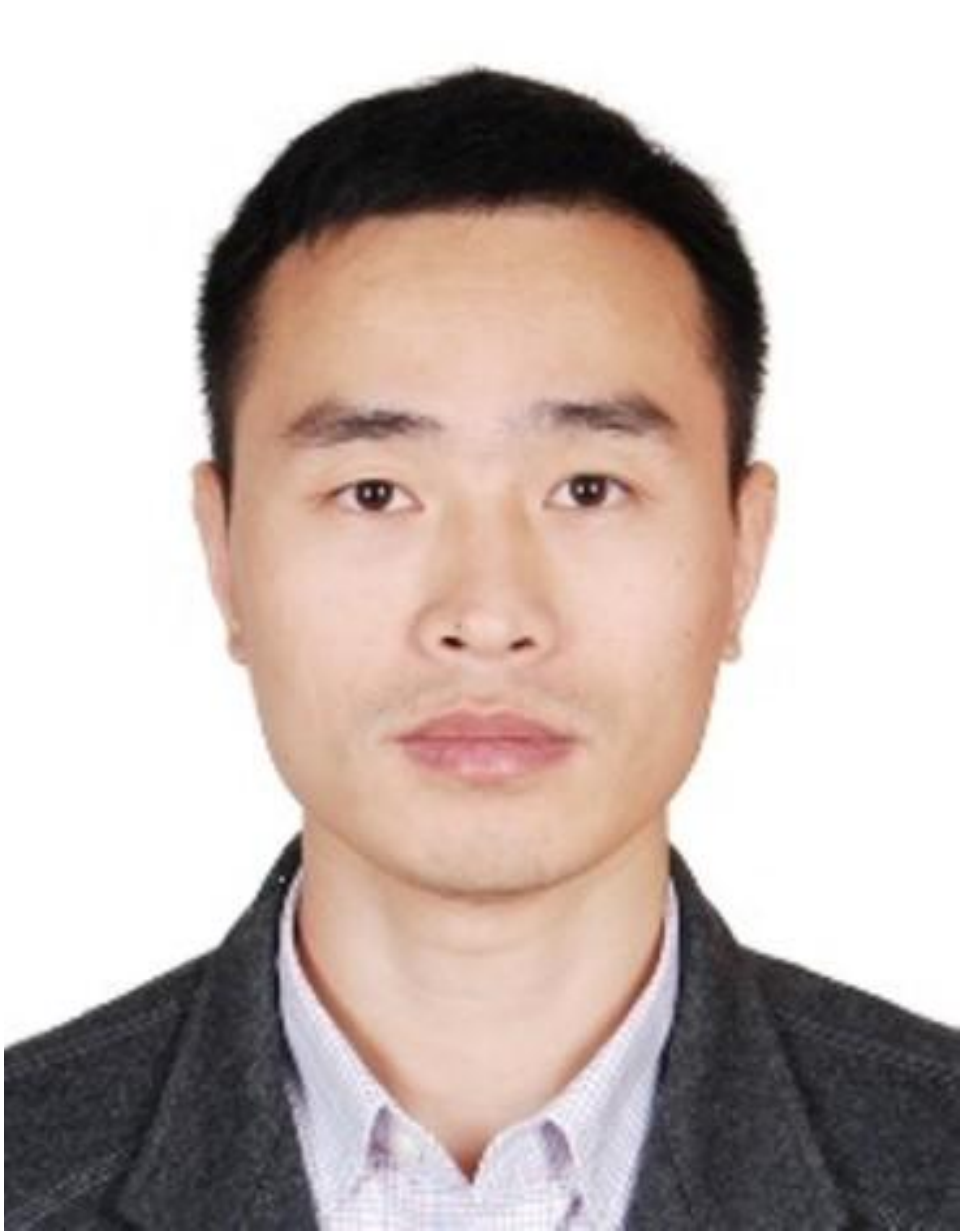}}]{Zhihua Xia} received his Ph.D. degree in computer science and technology from Hunan University, China, in 2011, and worked successively as a lecturer, an associate professor, and a professor with College of Computer and Software, Nanjing University of Information Science and Technology. He is currently a professor with the College of Cyber Security, Jinan University, China. He was a visiting scholar at New Jersey Institute of Technology, USA, in 2015, and was a visiting professor at Sungkyunkwan University, Korea, in 2016. He serves as a managing editor for IJAACS. His research interests include AI security, cloud computing security, and digital forensic. He is a member of the IEEE since Mar. 1, 2014.
\end{IEEEbiography}

\begin{IEEEbiography}[{\includegraphics[width=1in,height=1.25in,clip,keepaspectratio]{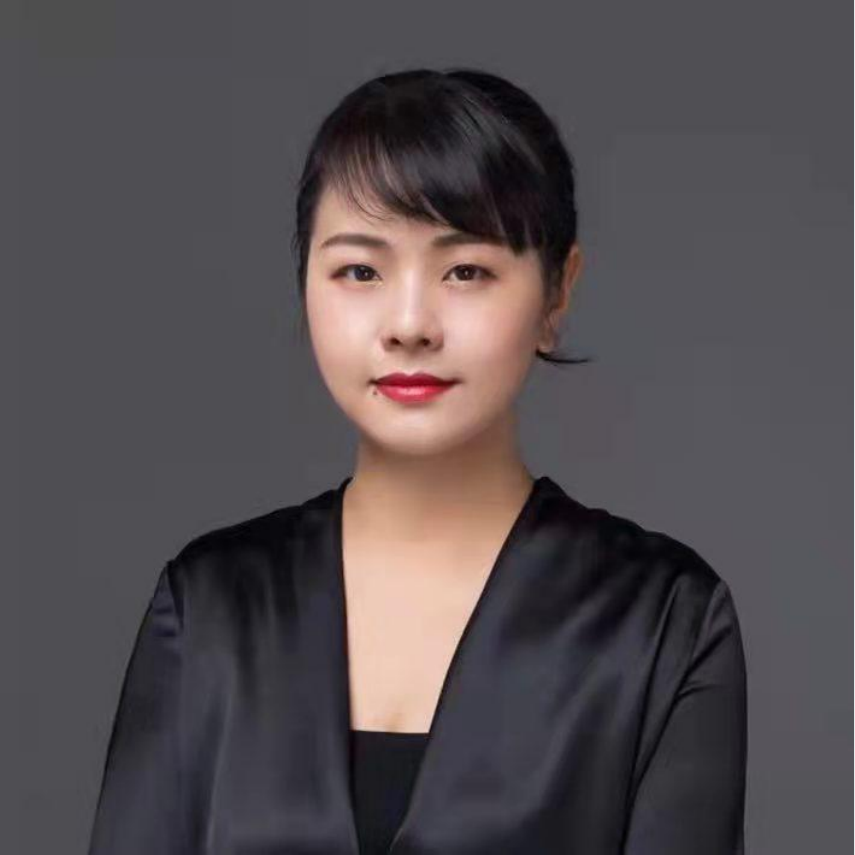}}]{Yao Tong} is an adjunct professor in the South China University of Technology, GuangZhou. Currently, she is the CEO of Guangzhou Fongwell Data Limited Company. Her research interests include data security, big data applications.
\end{IEEEbiography}

\begin{IEEEbiography}[{\includegraphics[width=1in,height=1.25in,clip,keepaspectratio]{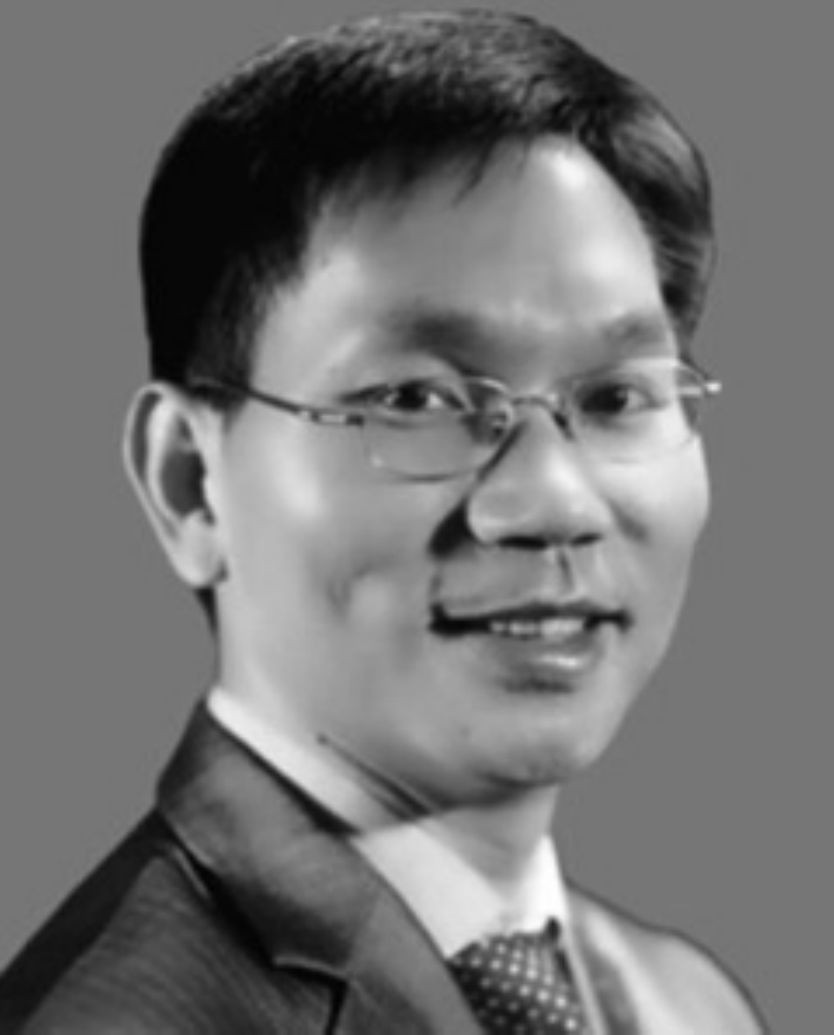}}]{Jian Weng} received the Ph.D. degree in computer science and engineering from Shanghai Jiao Tong University, Shanghai, China, in 2008. He is currently a Professor and the Dean with the College of Information Science and Technology, Jinan University, Guangzhou, China. His research interests include public key cryptography, cloud security, and blockchain. He was the PC Co-Chairs or PC Member for more than 30 international conferences. He also serves as an Associate Editor for the IEEE TRANSACTIONS ON VEHICULART ECHNOLOGY.
\end{IEEEbiography}

\begin{IEEEbiography}[{\includegraphics[width=1in,height=1.25in,clip,keepaspectratio]{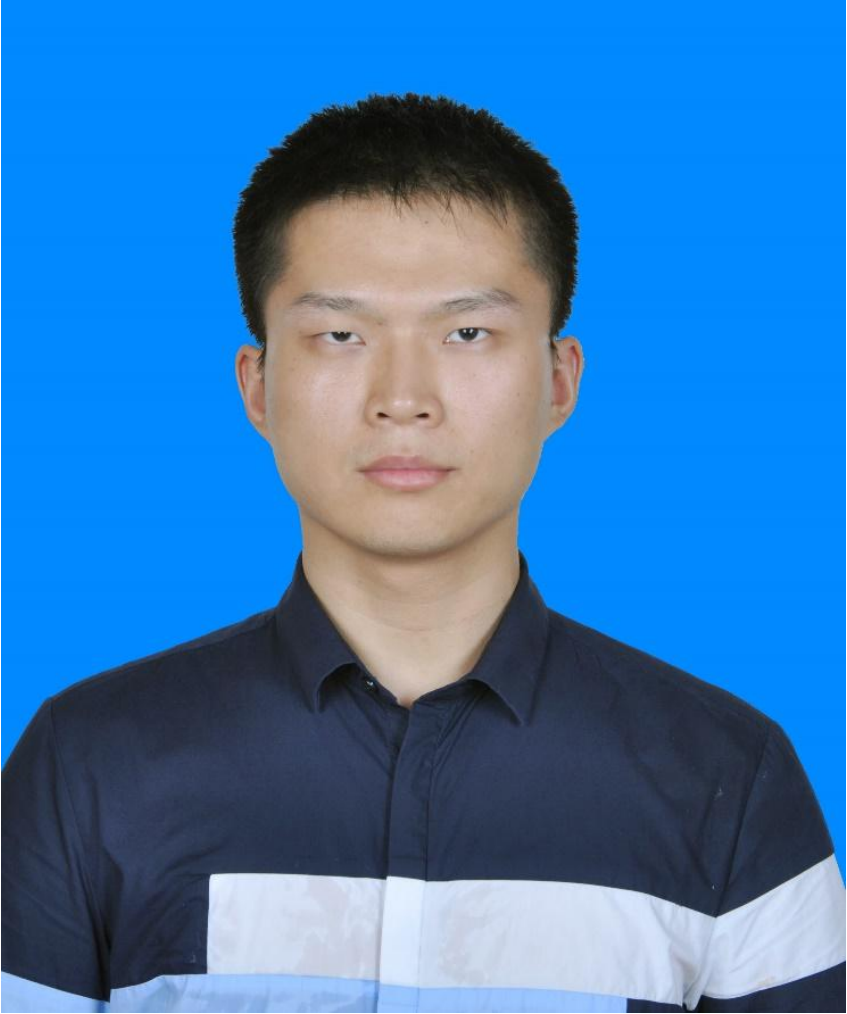}}]{Jianan Liu} (S’17) received the B.S. degree and M.S. degree from Zhengzhou University and Jinan University in 2013 and 2016 respectively. He is currently a Ph.D. candidate of Jinan University and a visiting scholar of Wilfrid Laurier University, Canada. His research interesting includes cryptography, smart grid security and cloud computing security.
\end{IEEEbiography}

\end{document}